\documentclass[journal,final]{IEEEtran}
\usepackage[scaled=1.0]{helvet}
\usepackage{amsmath}
\usepackage{amsfonts}
\usepackage{times}
\usepackage{graphicx}
\usepackage{subfigure}
\usepackage{color}
\usepackage{multirow}
\usepackage{parskip}
\usepackage{caption}
\usepackage{cite}

\begin{document}

\title{Probabilistic Spatial Distribution Prior Based Attentional Keypoints Matching Network}

\author{Xiaoming~Zhao,
	Jingmeng~Liu,
	Xingming~Wu,
	Weihai~Chen*,~\IEEEmembership{Member,~IEEE},	
	Fanghong~Guo,~\IEEEmembership{Member,~IEEE},
	and~Zhengguo~Li*,~\IEEEmembership{Senior~Member,~IEEE}% <-this % stops a space
	\thanks{Xiaoming Zhao, Jingmeng Liu, Xingming Wu, and Weihai Chen* are with the School of Automation Science and Electrical Engineering, Beihang University, Beijing, 100191 (e-mail: xmzhao@buaa.edu.cn, ljmbuaa110@163.com, wxmbuaa@163.com, and whchen@buaa.edu.cn).}% <-this % stops a space
	\thanks{Fanghong Guo is with the Department of Automation, Zhejiang University of Technology, Hangzhou 310014, China (email: fhguo@zjut.edu.cn).}% <-this % stops a space
	\thanks{Zhengguo Li* is with the SRO department, Institute for Infocomm Research, 1 Fusionopolis Way, Singapore (email: ezgli@i2r.a-star.edu.sg).}% <-this % stops a space
	\thanks{*\textit{(Corresponding author: Weihai Chen and Zhengguo Li.)}}}

%\markboth{IEEE TRANSACTIONS ON CIRCUITS AND SYSTEMS FOR VIDEO TECHNOLOGY,~Vol.~x, No.~x, xxx~xxxx}%
%{Shell \MakeLowercase{\textit{et al.}}: Bare Demo of IEEEtran.cls for IEEE Journals}

%\ninept
\maketitle

\begin{abstract}

Keypoints matching is a pivotal component for many image-relevant applications such as image stitching, visual simultaneous localization and mapping (SLAM), and so on. Both handcrafted-based and recently emerged deep learning-based keypoints matching methods merely rely on keypoints and local features, while losing sight of other available sensors such as inertial measurement unit (IMU) in the above applications. In this paper, we demonstrate that the motion estimation from IMU integration can be used to exploit the spatial distribution prior of keypoints between images. To this end, a probabilistic perspective of attention formulation is proposed to integrate the spatial distribution prior into the attentional graph neural network naturally. With the assistance of spatial distribution prior, the effort of the network for modeling the hidden features can be reduced. Furthermore, we present a projection loss for the proposed keypoints matching network, which gives a smooth edge between matching and un-matching keypoints. Image matching experiments on visual SLAM datasets indicate the effectiveness and efficiency of the presented method.

\end{abstract}


\IEEEpeerreviewmaketitle

\section{Introduction}

\IEEEPARstart{K}{eypoints} matching is an essential module in many image processing problems such as the visual SLAM, image stitching, and so on. It aims to establish 2D-2D matches (correspondences) of keypoints \cite{harris,shi,fast,keynet} between two images, so that the relative pose of cameras can be recovered with the multi-view geometry \cite{hartley_multiple_2003,ding_multi-camera_2020} or a set of differently exposed images \cite{wang_detail-enhanced_2020,zheng_single_2020}. Therefore, it becomes important to restore as many correct matches as possible. 

The matching of two point sets is a permutation problem. Matching $N$ points to other $N$ points leads to $N!$ possible permutations \cite{ma_locality_2019}. The standard approach to relieve this problem in image keypoint matching is to obtain a discriminative feature for each keypoint which is invariant to viewpoint, scale, illumination, and so on. And then the keypoint matches are recovered based on the similarity of features. A widely used heuristic strategy is firstly restoring a set of putative matches with the mutual nearest neighbor (mNN), and then filtering out false positive matches \cite{sift,pele_linear_2008,li_rejecting_2010} or regaining more consistent matches \cite{lipman_feature_2014,ma_robust_2014,lin_code_2017}. In addition to heuristic matching methods, several deep learning-based models \cite{pointcn,nmnet,oanet,sun_acne_2020} are proposed to exploit the local context based on the putative matches.

In this formulation, however, most local feature descriptors \cite{sift,surf,lift,matchnet,hardnet,sosnet} only encode an image patch with a limited region and ignore the global context. To track this problem, several end-to-end learned feature descriptors \cite{superpoint,d2net,r2d2} are proposed to extract global keypoint features implicitly. Nonetheless, the keypoints matching involves the geometric distribution and feature similarity of keypoints in/between images. Humans usually obtain this information by checking the feature points back-and-forth when matching keypoints between two images. This behavior is imitated with the self- and cross- attentional graph neural network (GNN) in the recently proposed SuperGlue \cite{superglue}. We extend this formulation with the observation that when humans have prior knowledge about the keypoints, they will compare and search for the matching keypoints based on the prior knowledge first, which will greatly reduce the matching effort. The search range is expanded only when no matching keypoints are found based on this prior knowledge. 

\begin{figure}[!tb]
%	\centering
 	\includegraphics[]{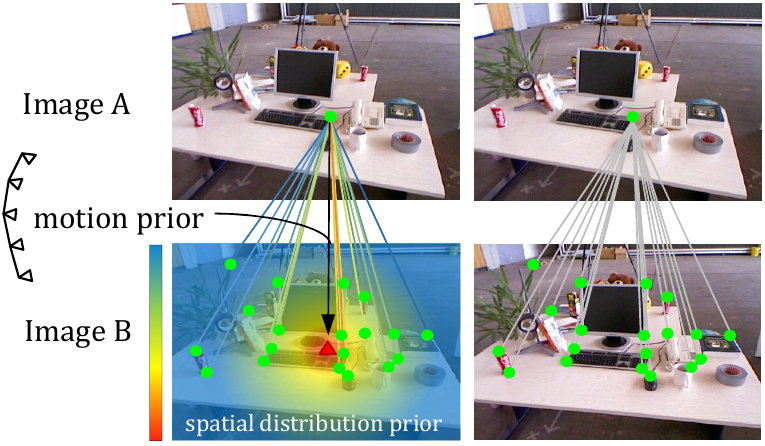}
	\caption{We propose to utilize the motion prior to exploit the spatial distribution prior of keypoints in two successive images, and the prior is used to assist the keypoints matching process. In this figure, a keypoint in image A is going to match with the keypoints in image B. In the left column, the proposed spatial distribution prior of the keypoint in image A is displayed superimposed on image B, which is used to assist the matching attentional graph neural network. While the right column shows the matching process without any prior.}
	\label{motivation}
\end{figure}

%Although of their remarkable performance, other knowledge that could assist and streamline the matching process is not fully considered.
As shown in the right column of Fig. \ref{motivation}, without prior knowledge, all the keypoints to be matched are treated equally without discrimination. So the SuperGlue \cite{superglue} has to refine the keypoint features with all contextual keypoints gradually through 18 attentional graph neural network (GNN) layers, which might be inefficient. On the other hand, the motion prior between two images could be obtained in SLAM systems using inertial measurement units (IMUs) or wheel odometry. For example, practical visual SLAM systems usually rely on a motion model to estimate an initial solution for the pose estimation between two consecutive images \cite{orbslam2,loam}. It is thus desired to assist the exploitation of keypoints geometric distribution and feature similarity cross image by using the motion prior, as shown in the left column of Fig. \ref{motivation}.

In this paper, we utilize the motion prior to compute the keypoint distribution prior, and present a keypoint distribution prior integration strategy of attentional GNN based on the probabilistic perspective of attention. More specifically, the initial pose of accurate IMU integration \cite{accimu,accimu2} is regraded as the motion prior of two successive images. Then, the spatial distribution prior of keypoints is obtained by warping keypoints across images with the motion prior. To integrate the spatial distribution prior into the attentional GNN, we propose to regard the attention as a Gaussian distribution that models the probability of the feature correlation. As such, the spatial distribution prior can be naturally and efficiently integrated into the attention module based on the conditional independence hypothesis. With the assistance of the prior, the attentional GNN can pay less effort to recover the correct matches. Thus, we streamline the network by decreasing some of the attentional layers. Furthermore, different from the matching loss, in which the ground-truth matches are obtained with a hard threshold of the keypoint projection errors, we propose to relax the hard threshold by directly utilizing the projection error in a margin.

The main contributions of this paper are in three folds:
\begin{itemize}
	\item With the probabilistic interpretation of attention, we exploit the motion prior from IMU measurements to obtain the spatial distribution prior of keypoints. Thus the keypoints matching network can be streamlined with fewer attentional GNN layers, resulting in less computational cost.
	\item We proposed to use the projection errors instead of the hard-threshold ground-truth matches to supervise the training of attentional GNN, so that the network can achieve better overall matching performance.
	\item Our approach can estimate keypoint matches efficiently and accurately on SLAM datasets such as InteriorNet \cite{interiornet_2018}, TUM-RGBD \cite{tumrgbd}, and ETH3D \cite{eth3d}.
\end{itemize}

The rest of this paper is organized as follows. Section \ref{related} reviews the handcrafted and deep learned keypoints matching methods for image pairs. Section \ref{method} begins by formulating the attentional GNN, introduces the direct and probabilistic spatial distribution prior integration to attention, and proposes the projection loss. In Section \ref{exp}, we evaluate and discuss the probabilistic prior integration and projection loss on different datasets. Finally, a conclusion is given in Section \ref{conclusion}.

\begin{figure*}[htb]
	\centering
	\includegraphics[]{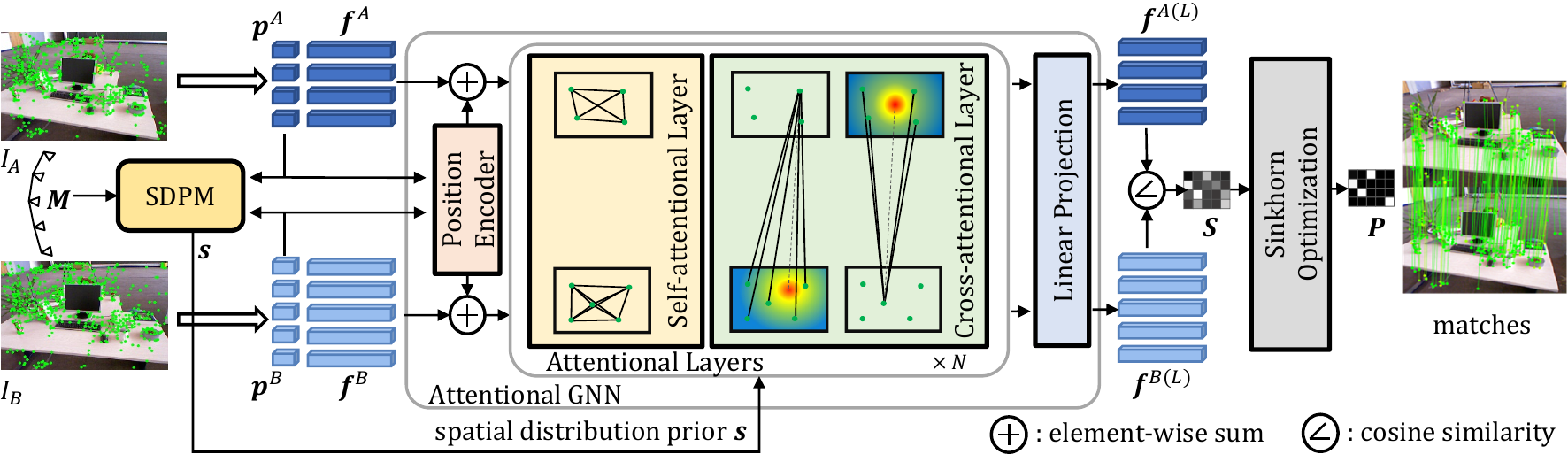}
	\caption{The proposed spatial distribution prior assisted matching pipeline. The $(\boldsymbol{p}^A, \boldsymbol{f}^A)$ and $(\boldsymbol{p}^B, \boldsymbol{f}^B)$ are local features extracted from two successive image frames $I_A$ and $I_B$ respectively, and $\boldsymbol M$ is the IMU measurements between $I_A$ and $I_B$. The \textbf{S}patial \textbf{D}istribution \textbf{P}rior \textbf{M}odule (SDPM) takes the $\boldsymbol{p}^A$, $\boldsymbol{p}^B$, and $\boldsymbol M$ as inputs and computes the spatial distribution prior $\boldsymbol{s}$. The attentional GNN then can utilize the spatial distribution prior $\boldsymbol{s}$ to assist the optimization of the keypoint features. At last, the Sinkorn algorithm is adopted to optimize the cosine similarity matrix $\boldsymbol S$ of the features, and it outputs the matching result $\boldsymbol P$.}
	\label{pipeline}
\end{figure*}
\section{Related works}
\label{related}
Keypoints matching for an image pair usually follows the following steps: a) detecting keypoints and extracting local features, b) finding the putative matches with brutally nearest neighbor matching, and then c) filtering out the false positive matches. 
For the first step, vast handcrafted \cite{harris,fast,sift,surf,orb} and learned \cite{matchnet,tfeat,lift,l2net,hardnet,lfnet,tilde,quadnet,keynet,superpoint,d2net,r2d2} methods have been proposed, devoting efforts to extract repeatable keypoints and discriminative local features. Whereas the latter two steps concentrate on retrieving accurate and exhaustive keypoint matches in an image pair based on the extracted keypoints and local features in the first step. They can also roughly be divided into two categories: the handcrafted and the learned.

\subsection{Handcrafted keypoints matching}

Researchers have designed massive approaches for keypoints matching based on heuristic experiences. The most successful and widely used one is the ratio test \cite{sift}, which is proposed along with the SIFT descriptor by Lowe. It removes the false positive matches by testing the similarity between the nearest and next-nearest neighbor. Subsequent research on ratio test improves the similarity measurement by using the  Earth Mover's Distance \cite{pele_linear_2008}. 

To recover more robust matches, various methods explores local keypoints distribution such as triangle constraint \cite{guo_good_2012}, vector field consensus \cite{ma_robust_2014}, local neighborhood structures \cite{ma_guided_2018,ma_locality_2019}, coherence-based separability constraint \cite{lin_code_2017}, and motion smoothness \cite{gms}. In addition to local keypoints distribution exploration, other strategies such as suitable local feature selection \cite{hu_matching_2015} and optical flow guided matching \cite{maier_guided_2016} have also been explored to recover more matches. However, the above methods could ignore the global consistency since they only focus on the local distribution.

On the other hand, the global properties of keypoints have also been explored. In such approaches, the matching problem is formulated as correspondence function \cite{li_rejecting_2010}, bounded distortion transformation \cite{lipman_feature_2014}, rigid transformation \cite{liu_regularization_2015}, or graph matching \cite{torresani_feature_2008,bbmatch,coor_gm,dcm}, and they can be solved iteratively. Our proposed method is based on graph neural network and is most similar to graph matching methods \cite{torresani_feature_2008,bbmatch,coor_gm,dcm} which are designed based on human experiences. For example, \cite{coor_gm} focuses on the second- and high-order graph matching and \cite{dcm} introduces a dual calibration strategy to model the correspondence relationship in points and edges respectively. Moreover, the RANSAC \cite{ransac} procedure in subsequent tasks could also be regarded as a false positive removing process, in which the potential matches are iteratively sampled to fit a model and the putative matches are classified as inliers and outliers based on the fitness to the model. However, the iterative solution of global formulations is less computational efficient, and the result could degrade when the underlying matching model differs from the predefined model.

\subsection{Learned keypoints matching}
In recent years, deep networks reveal their superior performance in various computer vision tasks including keypoints matching. The deep learned image keypoints matching can be roughly divided into two categories, i.e. semantic and geometric matching. The semantic keypoints matching approach matches the semantic keypoints in different instances of the same category of objects \cite{ufer_deep_2017,yu_hierarchical_2018,cur}, while the geometric matching matches the keypoints on different images of the same scene and does not consider high-level semantic information \cite{pointcn,nmnet,oanet,sun_acne_2020,superglue}. Both of them are robust to lighting conditions \cite{1kou2017,1zheng2013}. The keypoints matching existing in SLAM systems usually refers to the latter, thus we focus on the geometric keypoints matching in this paper.

The seminal work of the deep network for the point set is the PointNet \cite{pointnet}, in which the multi-layer perception (MLP) extracts the point feature and the max-pooling layer aggregates the global feature. Although the PointNet is originally invented for the classification and segmentation of unordered 3D points, it also encourages the design of putative matches filtering networks \cite{pointcn,nmnet,oanet,sun_acne_2020}. They firstly pair the putative keypoint matches as 4D quads. Then the networks take the 4D quads as input and output a score for each putative match. The false matches are filtered out with a threshold on the scores. Specifically, PointCN \cite{pointcn} aggregates the context of all matching features with Context-Normalization (CN) modules and estimates the essential matrix of an image pair with differentiable weighted eight points model. Following PointCN, NM-Net \cite{nmnet} mines compatibility-specific locality of keypoints to discover reliable local neighbors. And OA-Net \cite{oanet} introduces differentiable pooling and unpooling layers to exploit the global context. More recently, ACNe \cite{sun_acne_2020} extends the CN by introducing the local and global attentive weights. However, the putative matches filtering networks merely learn the keypoints geometric features, as they only take the keypoint positions as input and ignore the local features.

Recently, SuperGlue \cite{superglue} takes advantage of both keypoint positions and local features. It utilizes the attentional GNN to propagate and aggregate the contextual features both in intra-image and inter-image. In this way, each local feature in an image is refined with features in both images. Then the Sinkhorn \cite{sinkhorn} algorithm optimizes the matching score matrix and produces an assignment matrix representing the matching result. Our method is built on the architecture of attentional GNN introduced in \cite{superglue}. Besides the keypoint positions and local features, our method also utilizes the initial pose from IMU integration. The spatial distribution prior of keypoints is computed based on the initial pose to reduce the matching efforts.

\section{Method}
\label{method}
In this section, we first present a brief overview of the proposed pipeline. To incorporate with the prior, a spatial distribution prior module is then introduced. After formulating the attentional GNN, the direct and probabilistic spatial prior integration method of attentional GNN is proposed. Moreover, the matching loss is investigated and a more reasonable projection loss is presented.

\subsection{Pipeline overview}
\label{pipeline_overview}

\subsubsection{Problem formulation}
Formally, we consider two images $\{I_A, I_B\}$, their depth maps $\{\boldsymbol D^A,\boldsymbol D^B\}$ and the IMU measurements $\{\boldsymbol M_i = (\boldsymbol{\omega}_B,\boldsymbol{a}_B)_i | i\in[1,M]\}$ between $I_A$ and $I_B$, where $\boldsymbol{\omega}_B = (\omega_x,\omega_y,\omega_z)$ and $\boldsymbol{a}_B = (a_x,a_y,a_z)$ denote the angular velocity and linear acceleration in the IMU body frame respectively.
The keypoint features $\{\boldsymbol F_i = (\boldsymbol p_i,\boldsymbol f_i)|i\in[1,N]\}$ of each image are first extracted by the feature extractor, where $\boldsymbol p_i=(u,v)$ is the position of the i-th keypoint, $\boldsymbol f_i$ denotes the corresponding feature, and $N$ is the number of extracted keypoints in the image.
Assuming there are $N_A$ keypoints in $I_A$ and $N_B$ keypoints in $I_B$, thus the keypoint features of $\{I_A, I_B\}$ are represented as $\{\boldsymbol F^A_i\}_{i\in[1,N_A]}$ and $\{\boldsymbol F^B_i\}_{i\in[1,N_B]}$.
Our goal is to match the pixel coordinate $\{\boldsymbol p^A_i\}_{i\in[1,N_A]}$ and $\{\boldsymbol p^B_i\}_{i\in[1,N_A]}$, with the initial pose $\boldsymbol T_{AB}$ integrated from the IMU measurement $\{\boldsymbol M_i\}_{i\in[1,M]}$. Note that under this formulation, the keypoint features can be extracted with any existing feature extractor such as the handcrafted \cite{sift,surf,orb} or the learned \cite{hardnet,sosnet,superpoint,d2net,r2d2,lift} feature.

\subsubsection{Prior assisted attentional GNN matching pipeline}
The overview of the proposed matching pipeline is shown in Fig. \ref{pipeline}. First, the Spatial Distribution Prior Module (SDPM), which will be presented in detail in Section \ref{sdpm}, computes the spatial distribution prior $\boldsymbol{s}$ by using the keypoints position $\boldsymbol{p}^A$, $\boldsymbol{p}^B$, and the IMU measurements $\boldsymbol M$. Before the attentional GNN layers, the Keypoint Encoder encodes the keypoints position $\boldsymbol{p}$ into feature space $\boldsymbol{f}$. Then, the spatial distribution prior $\boldsymbol{s}$ is fed to the attentional GNN (Section \ref{attentional_gnn} and Section \ref{prior_assist}) to assist the keypoint features optimization. Each attentional GNN layer has two types of attention: self-attention and cross-attention. The spatial distribution prior $\boldsymbol{s}$ is integrated into these attention mechanisms in a probabilistic way. Next, the cosine similarity matrix $\boldsymbol S$, which is computed by the linear transformed features of the attentional GNN, is optimized by the Sinkhorn algorithm \cite{sinkhorn} (Section \ref{sinkhorn}) to obtain the assignment matrix $\boldsymbol P$ representing the matching result.

\begin{figure}[!tb]
	\centering
	\includegraphics[]{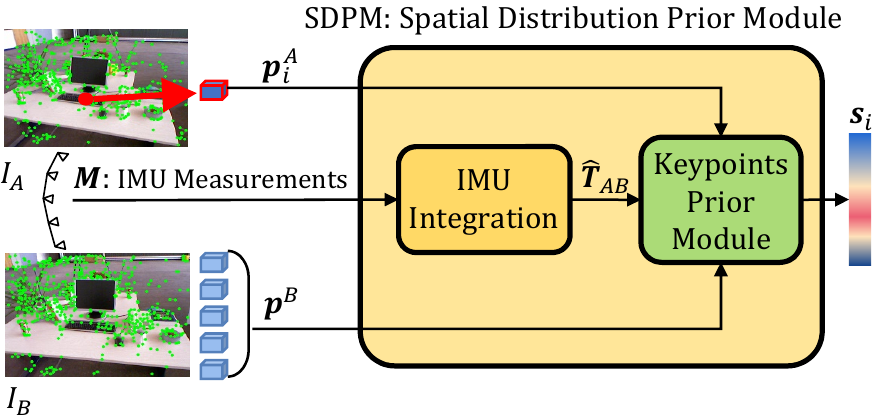}
	\caption{The Spatial Distribution Prior Module (SDPM). In this module, the IMU measurements $\boldsymbol{M}$ are first integrated into an initial pose $\hat{\boldsymbol{T}}_{AB}$. Then the Keypoints Prior Module computes the spatial distribution prior $\boldsymbol{s}$. For clarity, we take the $i$-th keypoint $\boldsymbol{p}_i^A$ in image $I_A$ as an example, and its spatial distribution prior $\boldsymbol{s}_i$ related to $\boldsymbol{p}^B$ is a normalized vector.}
	\label{spatial_prior}
\end{figure}

\subsection{Spatial distribution prior module}
\label{sdpm}
As shown in Fig. \ref{spatial_prior}, the spatial distribution prior module (SDPM) contains two sub-modules, the IMU integration and the keypoints prior module. The spatial distribution prior of the $i$-th keypoint in $I_A$ and all the keypoints in $I_B$ is illustrated for clarity.

\subsubsection{IMU Integration}
A basic motion assumption is to consider the velocity between two images is constant \cite{orbslam2, loam}. To obtain an accurate prior motion, IMU measurements have been widely used in recent SLAM systems to constrain the pose optimization problem \cite{viorb,vins, manifoldimu,accimu,accimu2}. In this paper, we adopt the switched linear system based IMU integration model \cite{accimu,accimu2}. It integrates all IMU measurements $\{\boldsymbol M_i = (\boldsymbol{\omega}_B,\boldsymbol{a}_B)_i | i\in[1,M]\}$ between $I_A$ and $I_B$ and outputs motion prior, that is, the initial relative pose $\hat{\boldsymbol{T}}_{AB}$ of $I_A$ and $I_B$.

\subsubsection{Keypoints Prior Module}
\label{kpm}
\begin{figure}[!tb]
	\centering
	\includegraphics[]{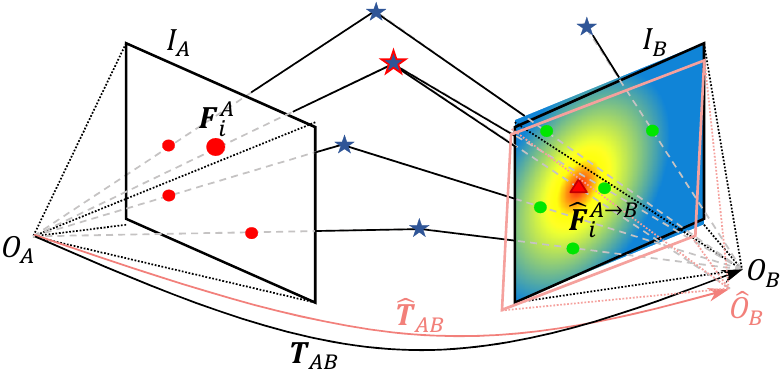}
	\caption{The keypoint prior module. The red and green dots represent the detected keypoints in image $I_A$ and $I_B$ respectively, and the blue pentacles denote the landmarks in the 3D world. For the $i$-th keypoints in $I_A$, this module warps it with the initial pose $\hat {\boldsymbol T}_{AB}$ into $I_B$ (the red triangle in $I_B$), and the spatial distribution prior $\boldsymbol{s}_i$ is formulated as a Gaussian distribution in $I_B$.}
	\label{warp}
\end{figure}

As illustrated in Fig. \ref{warp}, the keypoints prior module first warps the keypoint $\boldsymbol F_i^A=(\boldsymbol p_i,\boldsymbol f_i)^A$ in $I_A$ to $I_B$ with the motion prior $\hat {\boldsymbol T}_{AB}$:
\begin{equation}
	\hat {\boldsymbol p}_i^{A \rightarrow B} = \Pi(\hat {\boldsymbol T}_{AB}\Pi^{-1}(\boldsymbol p_i^A,\boldsymbol D^A)),
\end{equation}
where $\Pi(p)$ is the camera projection function that projects a 3D landmark $\boldsymbol l\in \mathbb{R}^3$ into the camera image plane.

We regard the warped position $\hat {\boldsymbol p}_i^{A \rightarrow B}$ in $I_B$ as keypoint prior knowledge across images. We name it as spatial distribution prior $s_{i,j}$ of keypoints and encode it with Gaussian distribution
\begin{equation}
	\label{s}
	s_{i,j} = \exp \frac{-(d_{ij}^{A,B})^2}{\sigma},	
\end{equation}
where $d_{ij}^{A,B}=\|\hat {\boldsymbol p}_i^{A \rightarrow B}-\boldsymbol p_j^B\|$ denotes the re-projection distances of keypoint $\hat {\boldsymbol p}_i^{A \rightarrow B}$ and $\boldsymbol p_j^B$.
Similarly, the spatial distribution prior of keypoints in the same image can be obtained with  $d_{ij}^{k,k}=\|\boldsymbol p_i^{k}-\boldsymbol p_j^k\| \ (k\in {A,B})$. In Section \ref{prior_assist}, we present two methods to integrate the spatial distribution prior ${\boldsymbol s}_i$ into attentional GNN.

\begin{figure*}[htb]
	\centering
	\includegraphics[]{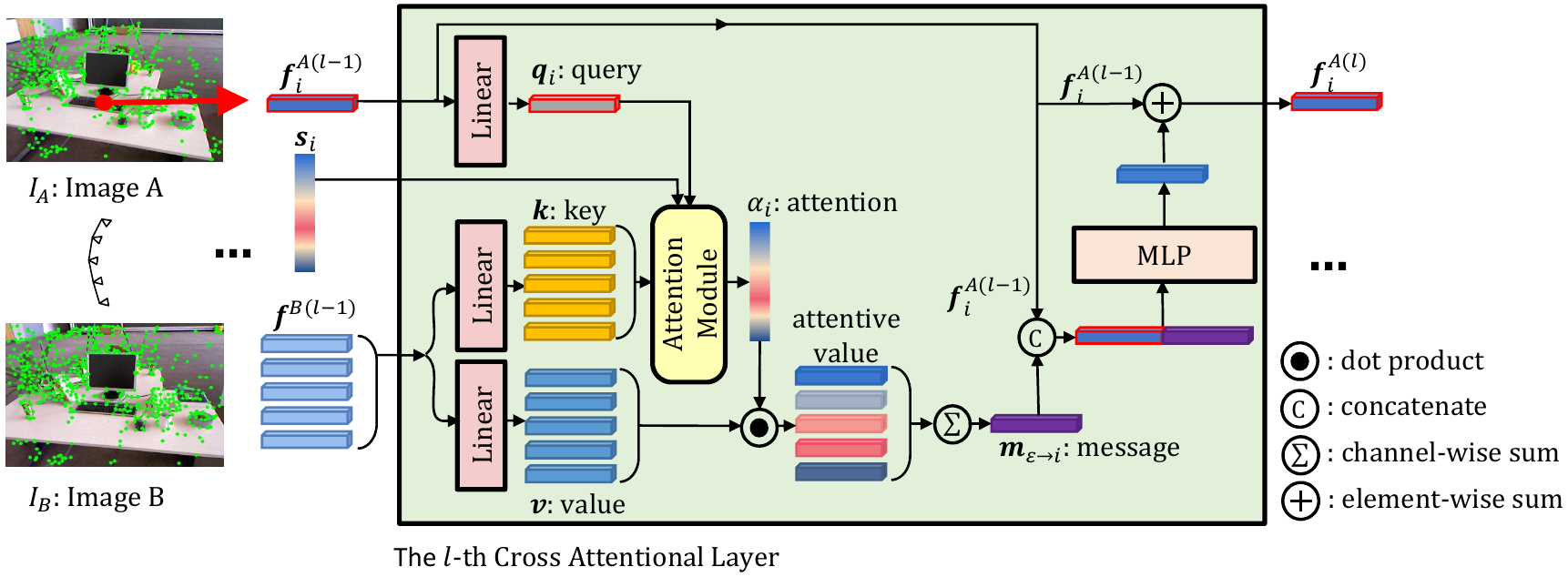}
	\caption{The cross attentional layer. The $i$-th feature $\boldsymbol f^{A(l-1)}_i$ in $I_A$ is optimized by all the features $\boldsymbol f^{B(l-1)}$ in $I_B$ and the spatial distribution prior in the $l$-th cross attentional GNN layer.}
	\label{gnn}
\end{figure*}

\subsection{Attentional GNN for keypoints matching}
\label{attentional_gnn}
The attentional GNN contains three modules, namely, Position Encoder, Attentional Layers, and the Linear Projection layer. 

\subsubsection{Position Encoder}
As shown in Fig. \ref{pipeline}, given the features $\boldsymbol F^A = \{\boldsymbol p^A,\boldsymbol f^A\}$ and $\boldsymbol F^B= \{\boldsymbol p^B,\boldsymbol f^B\}$ of an image pair, we first embed the keypoint position $\boldsymbol p$ into its feature space $\boldsymbol f$ with the position encoder. This enables the attentional GNN to utilize the position in an implicit way. Following \cite{superglue}, an MLP module is adopted as the position encoder for each feature $\boldsymbol F_i\in [1,N]$:
\begin{equation}
	\label{penc}
	\boldsymbol f_i^{(1)}=\boldsymbol f_i + \textup{MLP}(\boldsymbol p_i)\quad i\in [1,N].
\end{equation}

\subsubsection{Attentional Layers}
The attentional GNN layers for keypoints matching \cite{superglue} formulates the vertexes as keypoint feature $\boldsymbol f$, edges as attention $\alpha$. It optimizes each keypoint feature $\boldsymbol f$ iteratively to obtain the contextual information in-and-cross images. Fig. \ref{gnn} illustrates the process that the $i$-th feature in $I_A$ is optimized by all the features in $I_B$. In the following, to facilitate the notation, the superscripts $A$ and $B$ are omitted to denote arbitrary keypoint pairs of arbitrary images.

As in transformer \cite{atten_transformer}, the input features are first linear-transformed to the query $\boldsymbol{q}_i$, key $\boldsymbol{k}$, and value $\boldsymbol{v}$:
\begin{equation}
	\label{qkv}
	\begin{array}{lr}
		\boldsymbol{q}_i=\boldsymbol{W}_1 f_i^{(l-1)} + \boldsymbol{b}_1 \\ 
		\boldsymbol{k}_j=\boldsymbol{W}_2 f_j^{(l-1)} + \boldsymbol{b}_2 \\ 
		\boldsymbol{v}_j=\boldsymbol{W}_3 f_j^{(l-1)} + \boldsymbol{b}_3
	\end{array}
\end{equation}
where $j$ denotes the $j$-th feature in images. Then, the attention
\begin{equation}
	\label{atten_formula}
	\alpha_{ij}=\text{atten}(\boldsymbol{q}_i,\boldsymbol{k}_j,s_{i,j})
\end{equation}
is obtained to focus on different values $\boldsymbol{v}$. So that the contextual messages $\textup{m}_{\varepsilon\rightarrow i}$ for the feature $\boldsymbol f_i$ are propagated through all the edges $\varepsilon\rightarrow i$ connected with vertex $\boldsymbol f_i$:
\begin{equation}
	\label{message}
	\textup{m}_{\varepsilon\rightarrow i}=\sum_{j,(ij)\in\varepsilon}\alpha_{ij}\boldsymbol{v}_j,
\end{equation}

The message $\boldsymbol{m}_{\varepsilon\rightarrow i}$ is aggregated into the $i$-th feature $\boldsymbol f^{(l-1)}_i$ in a residual style. The residual is the MLP dimensionality reduction of the concatenated contextual messages $[\boldsymbol f_i^{(l-1)} \| \textup{m}_{\varepsilon\rightarrow i} ]$:
\begin{equation}
	\label{aggregation}
	\boldsymbol f_i^{(l)}=\boldsymbol f_i^{(l-1)}+\textup{MLP}([\boldsymbol f_i^{(l-1)}\| \textup{m}_{\varepsilon\rightarrow i}]).
\end{equation}

\subsubsection{Linear Projection}
In the above processes, the self- and cross- attentional GNN layers are used iteratively to imitate the back-and-forth behaviors when asking a human to match keypoints. For a specific feature $\boldsymbol f_i$ in one image, the self-attention GNN layer propagates contextual messages from the same image, while the cross-attention GNN layer obtains contextual messages from another image. After several self- and cross- attentional GNN layers, the final matching feature $\boldsymbol f_i$ are linear projections of the last attentional GNN layer outputs $\boldsymbol f_i^{(L)}$:
\begin{equation}
	\label{final}
	\boldsymbol f_i=\boldsymbol{W} \boldsymbol f_i ^{(L)} + \boldsymbol b.
\end{equation}

\subsection{Prior Assisted Attentional GNN for keypoints matching}
\label{prior_assist}

For the attention module in Fig. \ref{gnn}, no prior knowledge is considered in the vanilla formulation \cite{superglue}
{\small
\begin{equation}
	\label{vallina}
	\begin{aligned}
		\alpha_{ij} =\text{atten}(\boldsymbol{q}_i,\boldsymbol{k}_j,s_{i,j}) = \textup{softmax}_j(\boldsymbol{q}_i^T\boldsymbol{k}_j) = \frac{e^{\boldsymbol{q}_i^T\boldsymbol{k}_j}}{\sum_{j=1}^n e^{\boldsymbol{q}_i^T\boldsymbol{k}_j}},
	\end{aligned}	
\end{equation}
}
it has to learn to extract appropriate contextual cues from scratch. Thus multiple attentional GNN layers are necessary to enable the network to obtain correct context message gradually. This leads to more computational cost, which is not suitable for time-sensitive visual SLAM systems. To solve this problem, we propose to integrate the spatial distribution prior into the attention module.

\subsubsection{Direct spatial prior integration}
It is rational that a keypoint should have strong correlations with the ones close to it. Based on this intuition, the correlation strength of each keypoint should decrease with the distance to others. A straightforward approach to integrating the prior $s_{ij}$ into attentional GNN is using it to weigh the attention in Equation (\ref{message}). But directly weighting on $\alpha_{ij}$ will break the normalization property, thus we can weigh on the variable of softmax $\alpha_{ij} = \textup{softmax}_j(s_{ij}\boldsymbol{q}_i^T\boldsymbol{k}_j)$. However, when the distance $d_{ij}$ of feature $i$ and $j$ is greater than $3\sigma$, the prior $s_{ij}$ becomes very close to zero. Thus the attention on other hidden features will be weakened unexpectedly. So the direct spatial prior integration is formulated as
\begin{equation}
	\label{direct}
	\begin{aligned}
		\alpha_{ij} =\text{atten}(\boldsymbol{q}_i,\boldsymbol{k}_j,s_{i,j}) = \textup{softmax}_j((1+s_{ij})\boldsymbol{q}_i^T\boldsymbol{k}_j).
	\end{aligned}	
\end{equation}
It can be seen that the weight $(1+s_{ij})$ ranges from one to two, so it can preserve the attention on other hidden features even if $s_{ij}=0$. Under this formulation, the spatial prior could hinder the original attention extraction as they are tightly coupled together. So a probabilistic spatial prior integration is proposed to tackle this problem.

\subsubsection{Probabilistic spatial prior integration}

\begin{figure}[!tb]
	\centering
	\includegraphics[]{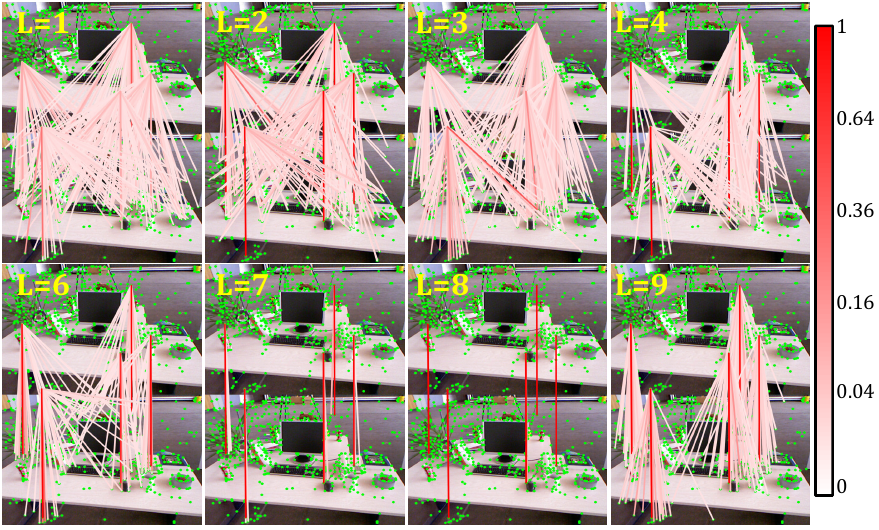}
	\caption{The visualization of the cross-attention of the vanilla attentional GNN layers in  \cite{superglue}. The cross-attention values of five keypoints in the first image are visualized, and the attentions lower than 0.01 are omitted for clarity. The first several vanilla attention layers focus on a large region, and it is not until the seventh layer that the network focus on the corresponding keypoints. Moreover, the ninth attention layer slightly expands its focus region.}
	\label{superglue_attn}
\end{figure}

To integrate the spatial prior into the attention module in a natural way, we exploit the probabilistic perspective of the attention.
In the formulation of the vanilla attention module, i.e. Equation (\ref{vallina}), the attention $\alpha_{ij}$ is a normalized Gaussian distribution \cite{wang2018non} of the cosine similarity of query $\boldsymbol q_i$ and key $\boldsymbol k_j$. With this perspective, the vanilla attention $\alpha_{ij}$ can be regarded as a joint distribution for each embedded features $\boldsymbol{q}_i$ and $\boldsymbol{k}_j$ as
\begin{equation}
	\Pr(\boldsymbol f_i,\boldsymbol f_j|\text{appearance},\text{spatial},...) = \alpha_{ij} \propto \exp(\boldsymbol{q}_i^T\boldsymbol{k}_j).
\end{equation}
In fact, it is a distribution conditioned by appearance, spatial characteristics, and other hidden contexts. Our intuition is that the vanilla attention module has to model all the hidden contexts from scratch, which could be difficult. This intuition can be further confirmed with the closer investigations on the attention $\alpha_{ij}$ in each layer. As shown in Fig. \ref{superglue_attn}, the network struggles to find a proper attentional relationship of features across images. With the assistance of spatial distribution prior, the attentional GNN can focus on the modeling of other contexts such as the appearance, and the attentional GNN layers can be streamlined to only two attentional GNN layers, as shown in Fig. \ref{prior_visualise}.

Then, we consider the spatial distribution prior $s_{ij}$ for keypoints $p_i$ and $p_j$ as a Gaussian distribution 
\begin{equation}
	\label{crossspatial}
	\Pr(\boldsymbol f_i,\boldsymbol f_j|\text{spatial}) \propto s_{ij} =  \exp(\frac{-d_{ij}^2}{\sigma}),
\end{equation}
and it is independent of other hidden contexts. As such, with the spatial distribution prior $\Pr(\boldsymbol f_i,\boldsymbol f_j|\text{spatial})$, the network can be relaxed to solely model the appearance and other hidden contexts $\Pr(\boldsymbol f_i,\boldsymbol f_j|\text{appearance},...)$. Thus the attentional distribution can be formulated as
\begin{equation}
	\label{int}
	\begin{aligned}
		\Pr(\boldsymbol f_i,\boldsymbol f_j|&\text{appearance},\text{spatial},...)  \\ & \propto  \Pr(\boldsymbol f_i,\boldsymbol f_j|\text{spatial})\Pr(\boldsymbol f_i,\boldsymbol f_j|\text{appearance},...) \\ & \propto  \exp(\frac{-d_{ij}^2}{\sigma})\exp(\boldsymbol{q}_i^T\boldsymbol{k}_j) \\ & \propto  \exp(\frac{-d_{ij}^2}{\sigma}+\boldsymbol{q}_i^T\boldsymbol{k}_j).
	\end{aligned}
\end{equation}

With the normalization of the above distribution, this process can be efficiently implemented as
\begin{equation}
	\label{prob_int}
	\begin{aligned}
	\Pr(\boldsymbol f_i,\boldsymbol f_j|\text{appearance},&\text{spatial},...) \\ &=\textup{softmax}(\frac{-d_{ij}^2}{\sigma}+\boldsymbol{q}_i^T\boldsymbol{k}_j).
\end{aligned}
\end{equation}

\subsection{Sinkhorn optimization}
\label{sinkhorn}

As shown in Fig. \ref{pipeline}, after the attentional GNN, a pairwise score matrix $\boldsymbol S = \{S_{ij}=(\boldsymbol{f}_i^A)^T\boldsymbol{f}_j^B \ | \ i\in [1,N_A], j \in [1,N_B]\}$ of features is first computed in the optimization layer. Following \cite{superglue}, two dustbins are added to the last column and row of $\boldsymbol S$ to form $\bar{\boldsymbol S}$ to cope with the unmatched keypoints. As the keypoint matching in image pair is a bipartite matching formulation in optimal transport \cite{optimal_transport}, its assignment matrix $\bar{\boldsymbol P}$ is given by
\begin{equation}
	\label{ot}
	L(\boldsymbol{a},\boldsymbol{b})=\min_{\boldsymbol P\in U(\boldsymbol{a},\boldsymbol{b})}\bar{\boldsymbol P}\odot\bar{\boldsymbol S}
\end{equation}
where $\odot$ denotes the Hadamard product operation, $\boldsymbol{a}=[\boldsymbol{1}_{N_A}^T\quad N_B]^T$ and $\boldsymbol{b}=[\boldsymbol{1}_{N_B}^T\quad N_A]^T$ are the mass of two discrete measures, $ U(\boldsymbol{a},\boldsymbol{b})$ denotes the assemble couplings under the constraints $\bar{\boldsymbol P}\boldsymbol{1}_{N_B+1}=\boldsymbol{a}$ and $\bar{\boldsymbol P}^T\boldsymbol{1}_{N_A+1}=\boldsymbol{b}$. The classical solution of bipartite matching is the Hungarian algorithm \cite{munkres_algorithms_1957}. Its differentiable version Sinkhorn algorithm \cite{sinkhorn} entropy-regularizes the soft assignment and can be solved on GPU efficiently.

The assignment matrix $\bar{\boldsymbol P}$ indicates the matching confidence of a keypoints pair. Then the matches are recovered by finding the row and column minimum and filtering out the false positives with a confidence threshold. 

\subsection{Loss functions}
\label{loss_funcs}
To train the proposed attentional GNN, we explore the matching loss proposed in \cite{superglue} and propose a projection loss in this section. The former utilizes hard-threshold matches while the latter uses projection errors to supervise the attentional GNN.
\subsubsection{Matching loss}

The elements in the assignment matrix $\bar{\boldsymbol P}_{1:N_{A},1:N_{B}}$ indicate the matching confidence for each possible keypoint pairs, and the elements in dustbins $\bar{\boldsymbol P}_{N_{A}+1,1:N_{B}}$ and $\bar{\boldsymbol P}_{1:N_{A},N_{B}+1}$ suggest the unmatched confidence. The direct way to constrain $\bar{\boldsymbol P}$ is by using the ground-truth keypoint matching $\mathcal{M} \subseteq \{(i,j)|i\in[1,N_A],j\in[1,N_B]\}$, unmatching $\mathcal{I}\subseteq [1,N_A]$, and $\mathcal{J}\subseteq [1,N_B]$.

To obtain the ground-truth matches $\mathcal{M}$, keypoints $\boldsymbol p_i^A$ in the image $I_A$ are warped to $I_B$ as $\boldsymbol p_i^{A\rightarrow B}$ with its depth map and ground-truth pose $\boldsymbol T_{AB}$. The distance matrix $\mathcal{D}\in \mathbb{R}^{N_A\times N_B}$ between the warped keypoints $\boldsymbol p_i^{A\rightarrow B}$ and $\boldsymbol p_j^B$ is computed. Then, the correct matching $\mathcal{M}$ is defined as
\begin{equation}
	\label{gtmatches}
	\mathcal{M} = \{(i,j)|d_{ij}<th,d_{ij}=\min d_{:j}, d_{ij}=\min d_{i:}, d_{ij} \in \mathcal{D}\},
\end{equation}
where $th$ is a pixel threshold.

As such, the matching loss could be formulated as the mean negative log-likelihood of the assignment $\bar{\boldsymbol P}$:
{\small
	\begin{equation}
		\label{pos_neg}
		\left\{
		\begin{aligned}
			&\mathcal{L}_{positive}=-\frac{\sum_{(ij)\in \mathcal{M}}\log \bar{\boldsymbol P}_{ij}}{\#\mathcal{M}}\\
			&\mathcal{L}_{negative}=-\frac{\sum_{i\in \mathcal{I}}\log \bar{\boldsymbol P}_{i,N_B+1} + \sum_{j\in\mathcal{J}}\log \bar{\boldsymbol P}_{N_A+1,j}}{\#\mathcal{I}+\#\mathcal{J}}\\
		\end{aligned}
		\right.,
\end{equation}}
where $\#\mathcal{M}$ denotes the total number of matched keypoint pairs, $\#\mathcal{I}$ and $\#\mathcal{J}$ are the number of unmatched keypoints in $I_A$ and $I_B$ respectively. Thus the total matching loss is given as
\begin{equation}
	\mathcal{L}_{total}= 2 \mathcal{L}_{positive} + \mathcal{L}_{negative}.
\end{equation}

\begin{figure}[!t]
	\centering
	\includegraphics[width=\hsize]{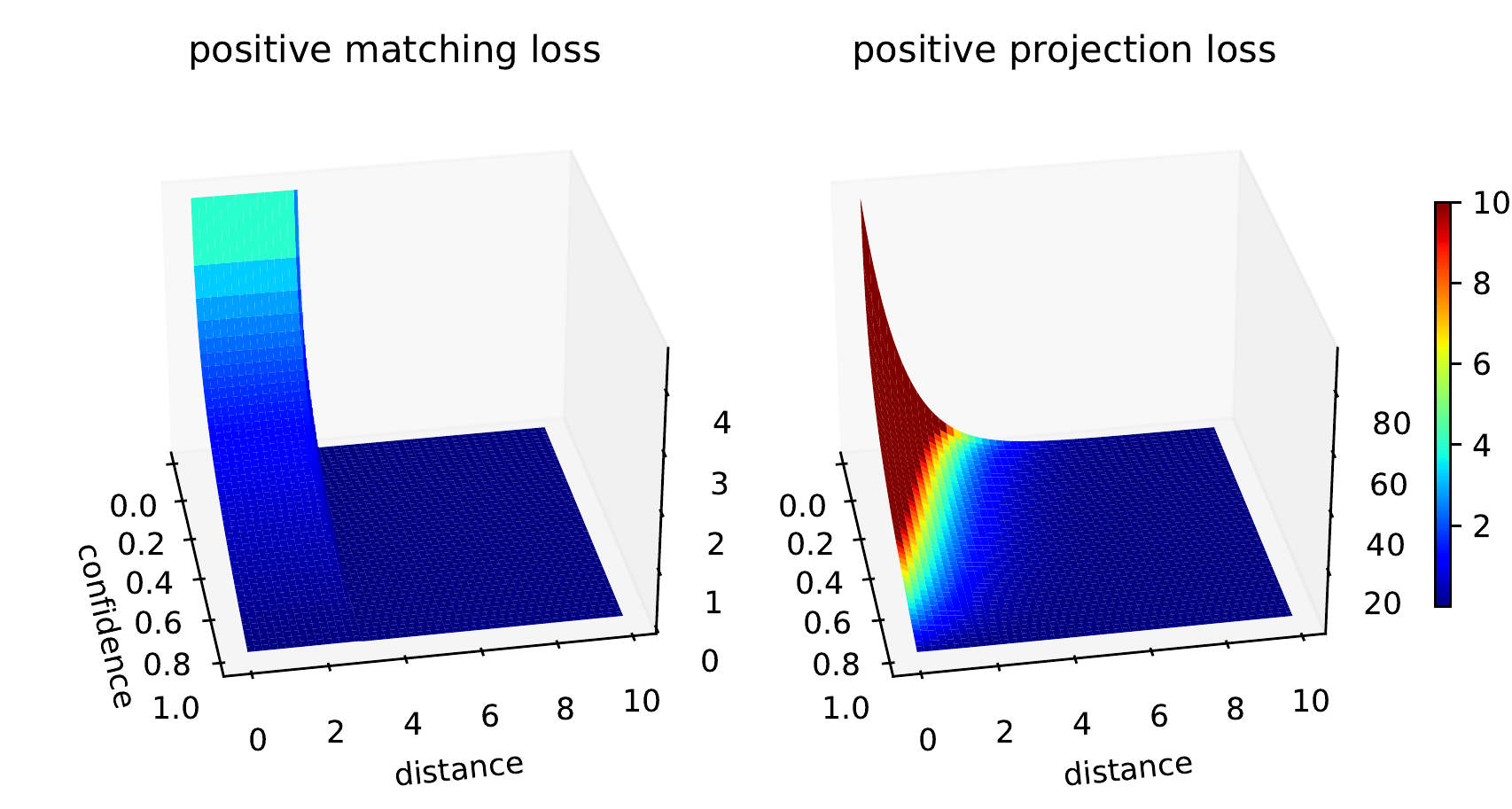}
	\caption{Positive losses  with respect to estimating matching confidence and ground-truth matching distance (with $th=3$ and $mg=10$). The matching loss truncates to zero when ground truth matching distance $d>th$, while the surface of the projection loss is much softer at the $th$.}
	\label{loss}
\end{figure}

\subsubsection{Projection loss}

However, in our experiments, we find that the model could not always converge to its best weights with the matching loss. We hypothesize that the hard-threshold ground-truth definition of Equation (\ref{gtmatches}) with a fix $th$ could be sub-optimal. As shown in Fig. \ref{loss}, the positive matching loss truncates the loss at $th$, regarding all the loss as the same when $d_{ij}<th$, while neglecting the loss for those $d_{ij}>th$.

To address this problem, we relax the threshold $th$ in Equation (\ref{gtmatches}) to a much larger margin $mg$ ($mg > th$), and the loss should be defined with the distance $d_{ij}$. To this end, a subset of $\mathcal{D}$ is defined as positive matches
\begin{equation}
	\label{D_sub}
	\tilde{\mathcal{D}} = \{d_{ij}|d_{ij}<mg,d_{ij}=\min d_{:j}, d_{ij}=\min d_{i:}, d_{ij} \in \mathcal{D}\}.
\end{equation}
And the unmatched keypoints $\mathcal{\tilde{I}}$ of $I_a$ and $\mathcal{\tilde{J}}$ of $I_B$ are those not in the positive matches $\tilde{\mathcal{D}}$.

Under this formulation, each element in $\tilde{\mathcal{D}}$ is a possible match and its value indicates the ground truth matching distance. Then the projection loss of all possible matches is given as
\begin{equation}
	\label{pos1}
	\mathcal{L}_{positive}=-\frac{\sum_{(ij)\in \tilde{\mathcal{D}}} \mathcal{L}_{projection}}{\#\tilde{\mathcal{D}}},
\end{equation}
where
\begin{equation}
	\label{projection}
	\mathcal{L}_{projection} = \exp(th-d_{ij}) \log \bar{\boldsymbol P}_{ij}.
\end{equation}

The projection loss can constrain the loss of positive samples softer and more reasonably, as shown in Fig. \ref{loss}. On the one hand, when the matching distance of keypoints is close to zero, they are more likely to be matched. In this case, the loss will push the output confidence to one. On the other hand, a larger ground truth matching distance indicates a weak tie, thus the loss constraint softly decreases with the distance. So that the loss can recall more potential matches.

\section{Experiments}
\label{exp}
In this section, we evaluated the proposed method on the one simulated and two real SLAM datasets. We first introduce experimental configurations including the datasets and implementation details. Then the proposed model is compared with state-of-the-art matching networks. To further inspect the proposed model, we discuss two types of prior integration and loss functions in the ablation studies.

\begin{figure*}[!ht]
	\centering
	\includegraphics[width=\linewidth]{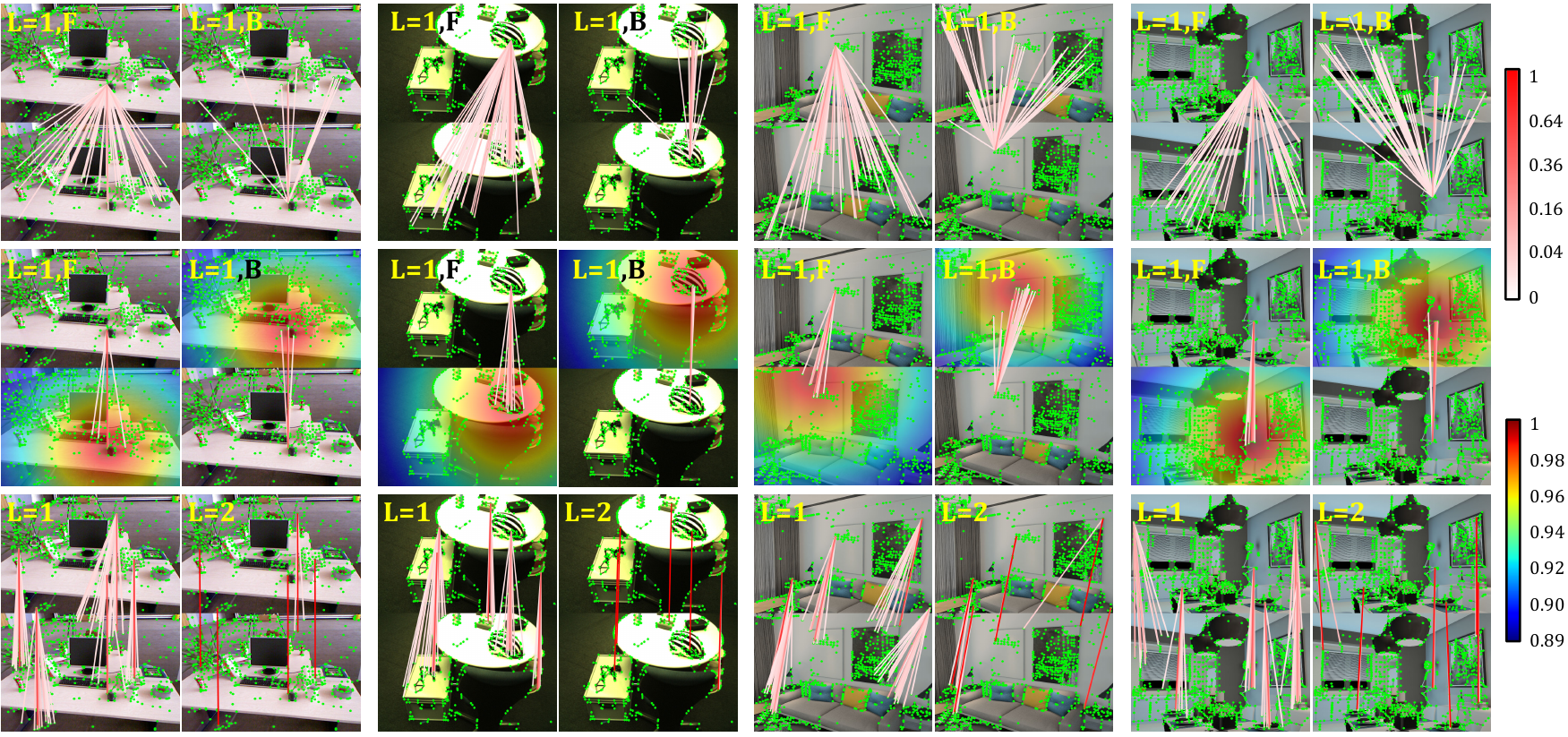}
	\caption{Visualization of the spatial distribution prior and its effectiveness. The image pairs are sampled from the TUM-RGBD \cite{tumrgbd}, ETH3D \cite{eth3d}, and the InteriorNet \cite{interiornet_2018} dataset. The cross-attention is drawn as lines across images, and the attentions lower than 0.01 are omitted for clarity. The ``L=x" represents layer number ``x", ``F" and ``B" denote the forward and backward cross-attention respectively. Top row: cross-attention of the first layer in SuperGlue \cite{superglue}; Middle row: cross-attention of the first layer in the proposed model, and the spatial distribution priors are overlapped on images; Bottom row: cross-attention visualization of different keypoints in the proposed model.  Notice that the prior assisted attentional GNN quickly can focus on the corresponding keypoints in the second attentional layer.}
	\label{prior_visualise}
\end{figure*}

\begin{figure}[!t]
	\centering
	\includegraphics[width=\hsize]{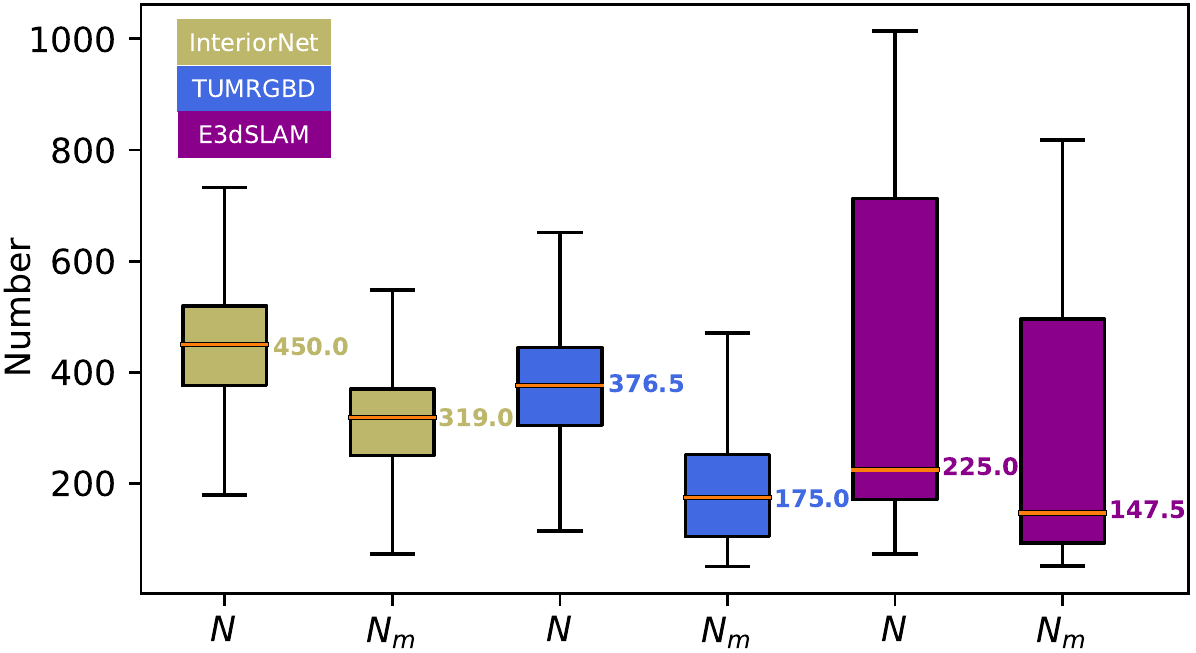}
	\caption{The keypoints and matches distribution of the test split for InteriorNet \cite{interiornet_2018}, TUM-RGBD \cite{tumrgbd}, and ETH3D \cite{eth3d} dataset. $N$ and $N_m$ denote the number of keypoints of images and the number of ground-truth matches of image pairs.}
	\label{dataset_s}
\end{figure}

\subsection{Datasets}
We train and evaluate the proposed model on three indoor SLAM datasets: InteriorNet \cite{interiornet_2018}, TUM-RGBD \cite{tumrgbd}, and ETH3D \cite{eth3d}. To check the generalization performance of the model, evaluations are also conducted on two homography datasets: HPatches \cite{hpatches_2017} and Oxford-Paris \cite{paris}. Detailed introductions of these datasets are as follows:

\textbf{InteriorNet dataset} \cite{interiornet_2018} provides 20 million synthetic RGBD images (RGB color images and corresponding depth maps), IMU data, and ground-truth camera poses. This dataset was created by professional designers based on real-world decorations. As it provides accurate depth maps and IMU measurements, there is no need to filter the data. We selected 21 trajectories for training, and 4 trajectories for testing.

\textbf{TUM-RGBD dataset} \cite{tumrgbd} contains the data captured by Kinect-V1 and motion capture system in three different indoor environments. The data includes well-calibrated RGBD images, ground truth camera poses, and accelerator readings for some sequences. As almost all the images were collected in a good light condition and rich textures, it is suitable for training the proposed matching model. To do so, we removed the sequences with dynamic objects and metallic spheres and selected 26 sequences for training, 7 sequences for testing.

\textbf{ETH3D dataset} \cite{eth3d} also provides the well-calibrated RGBD images and ground truth camera poses. Moreover, it includes time-aligned IMU measurements with linear acceleration and liner angular velocity for each sequence. Thus it is a proper training dataset for the proposed model. We selected the training and testing sequences from its training split. Those sequences with illumination changes, black color images, and dynamic objects are excluded. This ends up with 37 training sequences and 12 testing sequences.

\textbf{HPatches dataset} \cite{hpatches_2017} is a planar image pairs dataset with ground-truth homography matrix. It contains two subsets namely illumination and viewpoint, with 57 and 79 scenes respectively. This dataset is used to evaluate the generalization performance of the trained model.

\textbf{Oxford-Paris datatset} \cite{paris} has 6392 tourist pictures. We randomly selected 600 images and generated homography image pairs with the following producers. All the sampled images are first cropped and resized to size $640\times 480$. Then, a homography matrix is randomly sampled for each image in a way similar to \cite{superpoint}. At last, a corresponding image is generated with the homography transformation.

Since there is no motion prior for homography image pairs datasets, an initial homography matrix is obtained by adding noise to ground-truth homography transformation. As such, we can use the initial homography matrix to project the keypoints between images to attain the spatial distribution prior.

\begin{figure*}[!ht]
	\centering
	\includegraphics[]{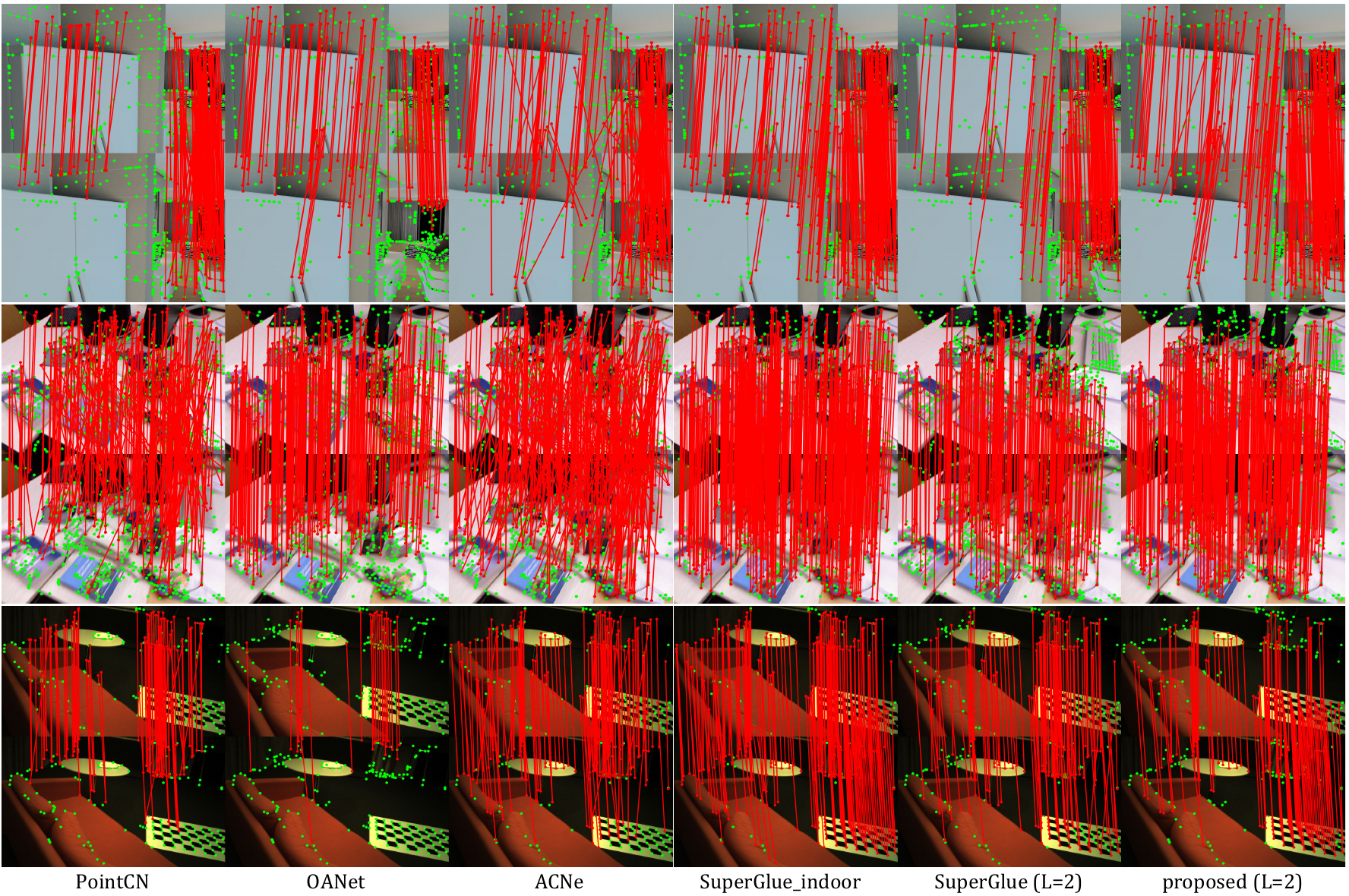}
	\caption{Matching examples of different learned methods on InteriorNet \cite{interiornet_2018} (top), TUM-RGBD \cite{tumrgbd} (middle), and ETH3D \cite{eth3d} (bottom) test set.}
	\label{images}
\end{figure*}

\begin{figure}[!ht]
	\centering
	\includegraphics[]{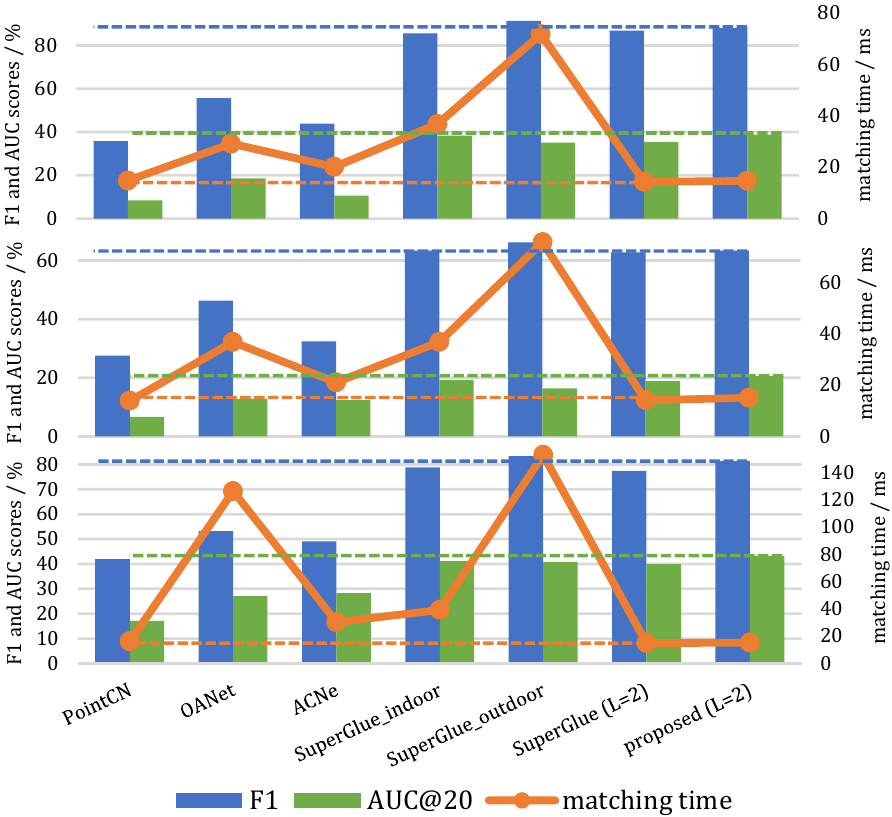}
	\caption{The F1 score, AUC@20\textdegree \ and the matching time of different learned methods on InteriorNet \cite{interiornet_2018} (top), TUM-RGBD \cite{tumrgbd} (middle), and ETH3D \cite{eth3d} (bottom) test set.}
	\label{f1auctime}
\end{figure}

\subsection{Implementation details}
\label{implementation}
To integrate the prior information to attentional GNN, we use single head attention rather than multi-head attention \cite{atten_transformer}. The position of keypoints is normalized by the image height and width. For direct prior integration, the $\sigma$ in Equation (\ref{s}) is fixed to 0.1. And for probabilistic prior integration, the $\sigma$ is a trainable parameter, such that the network can determine the spatial region of each layer. In all the experiments, the ground-truth matches are obtained with Equation (\ref{gtmatches}) with $th=3$.

To train the proposed models, we sample image pairs from InteriorNet \cite{interiornet_2018}, TUM-RGBD \cite{tumrgbd}, and ETH3D \cite{eth3d} dataset with a maximum of 300/300/200 image pairs, a maximum frame interval of 10/10/8, and a minimum overlap score of 0.3. The image pairs with less than 50 ground-truth matches are removed. An initial pose is integrated with the IMU measurements or accelerator readings between two sampled images \cite{accimu,accimu2}. If the translation error between the initial pose and the ground-truth pose is greater than 0.1 meters or the rotation error is greater than 8 degrees, then this image pair is discarded. For those image pairs in TUM-RGBD dataset \cite{tumrgbd} without valid IMU measurements, we generated a synthetic pose by sampling translation and rotation from a zero-mean Gaussian distribution with $\sigma=0.01\Delta t$.

During the training phase, we extracted 512 keypoints with SuperPoint \cite{superpoint} for each image. The non-maximum suppression radius is four pixels and keypoint threshold is 0.005. If there are no enough keypoints, additional keypoints were sampled from a uniform distribution to ensure that each image has 512 keypoints. To train the proposed models, we initialized them with the pre-trained position encoder and attentional GNN in SuperGlue \cite{superglue}. And the proposed models were trained on InteriorNet \cite{interiornet_2018}, TUM-RGBD \cite{tumrgbd}, and ETH3D \cite{eth3d} datasets together for a maximum of 300 epochs with adam optimizer \cite{adam} whose learning rate is $10^{-4}$ and batch size is 64. The validation loss was monitored, and if the validation loss was not reduced within 20 epochs, the training steps were early stopped. At last, the best models were selected and saved based on the validation loss.

While in the test and evaluation stages, the keypoints extraction configuration is the same as the training phase, except there is no limit on the number of keypoints. The distribution of keypoints and matches on the test set is shown in Fig. \ref{dataset_s}. The median numbers of keypoints are around 450, 377, and 225 for InteriorNet \cite{interiornet_2018}, TUM-RGBD \cite{tumrgbd}, and ETH3D \cite{eth3d}, respectively. The overall number of feature points ranges from 200 to 800, 100 to 700, and 80 to 1000 for the three datasets, respectively. Moreover, the number of ground truth matches of these three test sets is generally more than 100. The above conditions ensure that the image pairs of the test set have sufficient keypoints and matches, and can obtain accurate and comparable test results.

\begin{table*}[!t]		
	\centering 
	\caption{The matching performance of different methods on InteriorNet \cite{interiornet_2018}, TUM-RGBD \cite{tumrgbd}, and ETH3D \cite{eth3d} test sets. $P_m$, $R_m$, and $F1$ denote the matching precision, recall and F1 score respectively. L=2 and L=3 denote there are 2 and 3 self- and cross- attentional GNN layers in the model respectively. The best and the second best are marked as \textbf{bold} and \textcolor{blue}{\textbf{blue}}.}
	\begin{tabular}{c|ccc|ccc|ccc}
		\hline
		&                        \multicolumn{3}{c|}{InteriorNet \cite{interiornet_2018}}                        &                              \multicolumn{3}{c|}{TUM-RGBD \cite{tumrgbd}}                              &                                 \multicolumn{3}{c}{ETH3D \cite{eth3d}}                                 \\ \hline
		&              $P_m$               &              $R_m$               &               $F1$               &              $P_m$               &              $R_m$               &               $F1$               &              $P_m$               &              $R_m$               &               $F1$               \\ \hline
		mNN                 &              81.49               &              93.43               &              86.93               &              52.05               &              84.13               &              62.58               &              68.69               &              87.08               &              76.05               \\
		FLANN                &              82.26               &              87.96               &              84.90               &              52.52               &              76.71               &              60.61               &              69.02               &              76.36               &              71.51               \\
		PointCN \cite{pointcn}        &              52.36               &              27.70               &              35.77               &              33.14               &              25.68               &              27.55               &              53.91               &              35.87               &              41.90               \\
		OANet \cite{oanet}          &              78.74               &              44.62               &              55.63               &              53.86               &              46.71               &              46.33               &              67.91               &              48.16               &              53.13               \\
		ACNe \cite{sun_acne_2020}      &              48.16               &              40.41               &              43.83               &              30.86               &              36.78               &              32.38               &              50.16               &              49.26               &              49.01               \\
		SuperGlue\_indoor \cite{superglue}  &              78.89               &              94.06               &              85.65               &              52.08               &          \textbf{87.03}          & \textcolor{blue}{\textbf{63.37}} &              68.93               & \textcolor{blue}{\textbf{94.00}} &              78.76               \\
		SuperGlue\_outdoor \cite{superglue} &          \textbf{88.00}          &          \textbf{95.33}          &          \textbf{91.40}          & \textcolor{blue}{\textbf{56.53}} &              84.71               &          \textbf{66.15}          &          \textbf{75.27}          &          \textbf{95.06}          &          \textbf{83.37}          \\
		SuperGlue (L=2) \cite{superglue}   & \textcolor{blue}{\textbf{84.70}} &              89.43               &              86.83               &              55.10               &              77.01               &              62.79               &              71.21               &              85.90               &              77.39               \\
		proposed (L=2)            &              82.84               & \textcolor{blue}{\textbf{94.37}} & \textcolor{blue}{\textbf{88.10}} &          \textbf{60.98}          & \textcolor{blue}{\textbf{84.47}} &              63.32               & \textcolor{blue}{\textbf{72.94}} &              93.95               & \textcolor{blue}{\textbf{81.37}} \\ \hline
	\end{tabular}	
	\label{table:PRF}
\end{table*}

\begin{figure}[!t]
	\centering
	\includegraphics[]{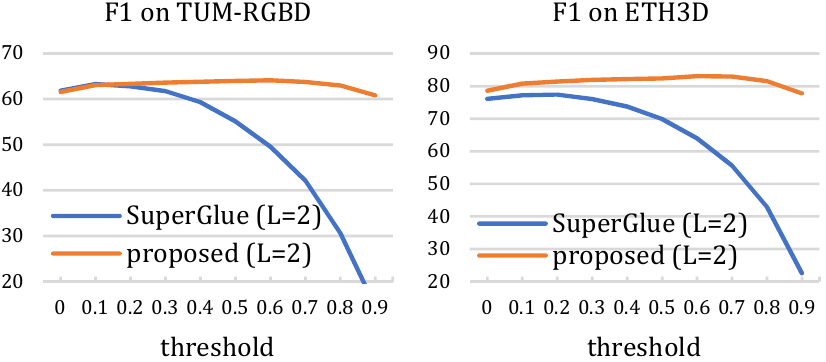}
	\caption{The F1 scores with respective to the different matching confidence thresholds of SueprGlue with two attentional GNN layers and the proposed method on TUM-RGBD \cite{tumrgbd} and ETH3D \cite{eth3d} test sets.}
	\label{matching_result_f1}
\end{figure}

\subsection{Visualization of the spatial distribution prior}

The visualization of the cross-attention and spatial distribution prior is shown in Fig. \ref{superglue_attn} and Fig. \ref{prior_visualise}. Without the assistance of any prior information, the SuperGlue \cite{superglue} has to model the attention from scratch. The model has to focus on almost all the keypoints in another image in the first few layers to find a potential match for one keypoint (Fig. \ref{superglue_attn} and the top row of Fig. \ref{prior_visualise}). And the proposed spatial distribution prior of keypoints across images is shown in the middle row of Fig. \ref{prior_visualise}. We can see that the spatial distribution prior gives a region of potential matches in another image. With the assistance of spatial distribution prior, the proposed model can focus on the potential keypoints in the first cross-attention layer (middle row of Fig. \ref{prior_visualise}), and quickly focus on the correct matches in the second cross-attention layer (bottom row of Fig. \ref{prior_visualise}). 

\subsection{Comparison to related works}

\subsubsection{Setups}
We compared the proposed method with the handcrafted and learned methods. The proposed method of two attentional GNN layers was trained with projection loss (will be discussed in Section \ref{ablation_sec}) for comparisons. For the handcrafted methods, two variations of nearest neighbor (NN) matching, the mutual NN with PyTorch implementation, and the FLANN with OpenCV implementation when the test ratio is 0.7,  were evaluated. For the learned methods, the PointCN \cite{pointcn}, OANet \cite{oanet}, ACNe \cite{sun_acne_2020}, and SuperGlue \cite{superglue} were assessed. During testing, the official released code and model weights were used. Specifically, we used the model trained on GL3D-v2 for OANet \cite{oanet}. For ACNe \cite{sun_acne_2020}, the indoor model weights were used, and the keypoint matches with a combined weight is greater than $10^{-7}$ were taken as correct matches. For SuperGlue \cite{superglue}, we assessed the official ``indoor" and ``outdoor" model weights, as well as the SueprGlue of two attentional GNN layers trained on our training set with the same training configures as for the proposed methods. The matching confidence thresholds of the assignment matrix of SuperGlue and the proposed models were both set to 0.2 as in \cite{superglue}, and we found this configuration gives the best trade-off between precision and recall. All the models were evaluated on a desktop platform with Intel Core i5-4590 and GeForce GTX TITAN X (PASCAL).

\subsubsection{Qualitative comparisons}
The matching results of different learned methods are shown in Fig. \ref{images} and Fig. \ref{f1auctime}. As can be seen from these figures, the PointNet-like methods generate some false matches (results of PointCN \cite{pointcn} and ACNe \cite{sun_acne_2020} in Fig. \ref{images}) and obtain fewer matches (results of PointCN \cite{pointcn} and OANet \cite{oanet} in Fig. \ref{images}) than the feature-based models. The possible reason is that the PointNet-like methods only take advantage of the position of keypoints and cannot model the appearance or other hidden features. The full SuperGlue \cite{superglue} model, on the other hand, achieves the best matching results on the test images. Nevertheless, the full SuperGlue model consumes a lot of computation time (Fig. \ref{f1auctime}), and is not suitable for real-time SLAM systems. Simplifying SuperGlue by reducing the number of GNN layers can decrease its computational efforts, but it also degrades the matching performance (as shown in Fig. \ref{f1auctime}). With the assistance of prior information, the proposed method fills this performance decline without increasing the computation time (as shown in Fig. \ref{f1auctime}).

\begin{table*}[!t]		
	\centering 
	\caption{The pose estimation AUC under 5\textdegree, 10\textdegree \ and 20\textdegree \ of the learned methods on InteriorNet \cite{interiornet_2018}, TUM-RGBD \cite{tumrgbd} and ETH3D \cite{eth3d} test sets. The best and the second best are marked as \textbf{bold} and \textcolor{blue}{\textbf{blue}}.}
	\begin{tabular}{c|rrr|rrr|rrr}
		\hline
		                                    &      \multicolumn{3}{c|}{InteriorNet \cite{interiornet_2018}}      &                             \multicolumn{3}{c|}{TUM-RGBD \cite{tumrgbd}}                             &                                 \multicolumn{3}{c}{ETH3D \cite{eth3d}}                                 \\ \hline
		                                    &          AUC@5 &         AUC@10 &                           AUC@20 &                           AUC@5 &                          AUC@10 &                           AUC@20 &                            AUC@5 &                           AUC@10 &                           AUC@20 \\ \hline
		      PointCN \cite{pointcn}        &           2.18 &           3.89 &                             8.28 &                            1.62 &                            3.43 &                             6.67 &                             2.37 &                             7.25 &                            17.15 \\
		        OANet \cite{oanet}          &           3.48 &           8.39 &                            18.52 &                            1.43 &                            4.26 &                            12.75 &                             7.14 &                            14.93 &                            27.04 \\
		     ACNe \cite{sun_acne_2020}      &           4.69 &           6.46 &                            10.52 & \textcolor{blue}{\textbf{4.49}} &                            7.29 &                            12.46 &                             8.24 &                            15.26 &                            28.29 \\
		SuperGlue\_indoor \cite{superglue}  & \textbf{13.72} & \textbf{24.01} & \textcolor{blue}{\textbf{38.26}} &                            3.19 &                            9.12 & \textcolor{blue}{\textbf{19.29}} & \textcolor{blue}{\textbf{11.03}} & \textcolor{blue}{\textbf{23.40}} & \textcolor{blue}{\textbf{41.04}} \\
		SuperGlue\_outdoor \cite{superglue} &          \textcolor{blue}{\textbf{10.27}} &          20.43 &                            34.99 &                            3.74 &                            6.86 &                            16.37 &                             9.33 &                            22.41 &                            40.85 \\
		 SuperGlue (L=2) \cite{superglue}   &           9.86 &          19.36 &                            35.37 &                            3.45 & \textcolor{blue}{\textbf{9.42}} &                            18.90 &                             6.97 &                            22.98 &                            39.94 \\
		          proposed (L=2)            &           9.91 &          \textcolor{blue}{\textbf{21.76}} &                   \textbf{38.80} &                   \textbf{6.37} &                  \textbf{11.35} &                   \textbf{20.49} &                   \textbf{11.47} &                   \textbf{25.53} &                   \textbf{43.23} \\ \hline
	\end{tabular}	
	\label{table:auc}
\end{table*}

\subsubsection{Matching results}
The ground-truth matches are first obtained according to Equation (\ref{gtmatches}) with $th=3$. Then the matching precision $P_m$, recall $R_m$, and F1-score is given as
\begin{equation}
	\begin{array}{lr}
		P_m=TP / (TP+FP)\\
		R_m=TP / (TP+FN)\\
		F1=2P_mR_m / (P_m+R_m)
	\end{array}
\end{equation}
where $TP$, $FP$, and $FN$ denote the true positives, false positives, and false negatives respectively. The $TP$, $TP+FP$, and $TP+FN$ represent the number of correct matches, estimated matches, and ground-truth matches in the keypoints matching case respectively.

\begin{figure}[!tb]
	\centering
	\includegraphics[width=\linewidth]{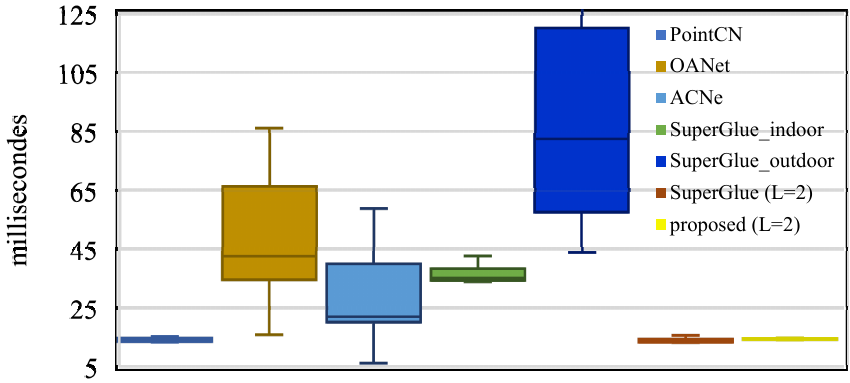}
	\caption{The running time distribution of the learned methods on ETH3D \cite{eth3d} test set. The outliers are not shown in this figure for clarity.}
	\label{time}
\end{figure}

The detailed matching results are shown in Table \ref{table:PRF}. It is noticeable that the mNN and FLANN both have good matching precision ($P_m$) and recall ($R_m$) on our test sets, which is only slightly lower than the SuperGlue \cite{superglue} and the proposed method. While the PointNet-like models PointCN \cite{pointcn}, OANet \cite{oanet}, and ACNe \cite{sun_acne_2020} do not perform well on our test sets. This could due to the fact that they follow the putative filtering strategy, so they require massive putative keypoint pairs (usually $>2000$) for each image pair. However, we only extracted the most distinctive keypoints as in most SLAM systems, resulting in fewer keypoints in the image and fewer putative matches in the image pairs. The full SuperGlue \cite{superglue} model with ``outdoor" weights (nine attentional GNN layers) achieves the best matching F1 scores, with an F1 score of 91.40\%, 66.15\%, and 83.37\% on InteriorNet \cite{interiornet_2018}, TUM-RGBD \cite{tumrgbd}, and ETH3D \cite{eth3d} respectively. Nonetheless, the SuperGlue of two attentional GNN layers drop the matching performance by 4.57\%, 3.36\%, and 5.98\% for the F1 score on InteriorNet, TUM-RGBD, and ETH3D test sets respectively. The proposed methods, which are assisted by the spatial distribution prior of keypoints, can regain the performance drop of reduction of GNN layers as in Table \ref{table:PRF}. Specifically, the proposed method promotes the recall $R_m$ by 4.94\%, 7.46\%, and 8.05\%, F1 scores by 1.27\%, 0.53\%, and 3.98\% on InteriorNet, TUM-RGBD, and ETH3D test sets receptively. Furthermore, as shown in Fig. \ref{matching_result_f1}, without the assistance of the spatial distribution prior, SuperGlue of two attentional GNN layers fails to recover correct matches with higher matching confidence (lower F1 score than the proposed method). Some matching examples are shown in Fig. \ref{images}.

\subsubsection{Pose estimation accuracy}

The purpose of keypoints matching is to estimate the relative pose of two images. So the pose accuracy is also evaluated. Given the predicted keypoint matches of an image pair, the essential matrix is obtained with OpenCV function \texttt{findEssentialMat}, and then the relative pose is recovered with \texttt{recoverPose}. As in previous works \cite{pointcn,oanet,superglue}, we calculate the pose angular differences between ground truth and estimated pose, and report the area under curve (AUC) with a maximum error of threshold 5\textdegree, 10\textdegree, and 20\textdegree.

\begin{table}[!t]		
	\centering 
	\caption{The matching precision $P_m$, recall $R_m$, $F1$ scores, and homography accuracy $Acc_H$ on homographic datasets. HPatches\_i and HPatches\_v denote the illumination and viewpoint subset of HPatches \cite{hpatches_2017}. The OxfordParis is the manually generated homography image pairs from OxfordParis dataset \cite{paris} as described in Section \ref{implementation}. The matching confidence of the model is 0.2.}
	\begin{tabular}{cccccc}
		\hline
		Dataset   &       model        & $P_m$ & $R_m$ & $F1$  & $Acc_H$  \\ \hline
		& SuperGlue\_outdoor & 80.75 & 84.92 & 82.52 & 94.39 \\
		HPatches\_i &  SuperGlue (L=2)   & 61.82 & 60.92 & 60.77 & 94.74 \\
		&   proposed (L=2)   & 62.42 & 67.89 & 64.73 & 92.63 \\ \hline
		& SuperGlue\_outdoor & 82.08 & 83.67 & 82.73 & 56.95 \\
		HPatches\_v &  SuperGlue (L=2)   & 55.64 & 51.23 & 52.13 & 45.76 \\
		&   proposed (L=2)   & 61.01 & 63.73 & 62.18 & 50.17 \\ \hline
		& SuperGlue\_outdoor & 76.12 & 76.54 & 75.51 & 76.50 \\
		OxfordParis &  SuperGlue (L=2)   & 52.49 & 42.77 & 45.58 & 55.33 \\
		&   proposed (L=2)   & 52.26 & 52.73 & 51.42 & 61.83 \\ \hline
	\end{tabular}	
	\label{table:homo}
\end{table}

Five learned models are assessed in Table \ref{table:auc}. The Pointnet-like models PointCN \cite{pointcn}, OANet \cite{oanet}, and ACNe \cite{sun_acne_2020} yield lower AUCs. The possible reason is that the number of keypoints is not sufficient for them to recover good matches. The visualization in Fig. \ref{images} suggests they recall fewer matches and produces many false matches. The AUCs of the full SuperGlue \cite{superglue} model, are generally higher than the PointNet-like models. And the SuperGlue with ``indoor" weights obtains the best AUC performance among previously learned methods, yields 38.26\%, 19.29\%, and 41.04\% of AUC@20\textdegree \ on InteriorNet \cite{interiornet_2018}, TUM-RGBD \cite{tumrgbd}, and ETH3D \cite{eth3d} test sets respectively. As the SuperGlue of two attentional GNN layers was trained on our training set, it also gives AUCs on par with the full SuperGlue model, with 35.37\%, 18.09\%, and 39.94\% of AUC@20\textdegree \ on InteriorNet, TUM-RGBD, and ETH3D test sets respectively. Due to the assistance of spatial distribution prior, the proposed method produces higher AUCs than SuperGlue of two attentional GNN layers by 3.43\%, 1.59\%, and 3.29\% of AUC@20\textdegree \ on InteriorNet, TUM-RGBD, and ETH3D test sets respectively, and outperforms the other learned methods.

\subsubsection{Running time}

We tested all the learned models with the same agenda to measure the matching time of an image pair. The results are shown in Fig. \ref{time}. Among the learned matching networks, the model sizes of ACNe \cite{sun_acne_2020}, OANet \cite{oanet}, and SuperGlue \cite{superglue} are larger, which results in longer running times. Since the running time is influenced by the number of points, the larger model in turn amplifies the dispersion of the running time. While the PointCN \cite{pointcn}, SuperGlue \cite{superglue} of two GNN layers and the proposed method have the shortest matching time of about 15 milliseconds. Considering the matching metrics and pose estimation accuracy of the proposed method are much better than those of PointCN \cite{pointcn} and SuperGlue \cite{superglue} of two GNN layers, the effectiveness of the proposed prior integration and projection loss is verified.

\begin{figure}[!tb]
	\centering
	\includegraphics[]{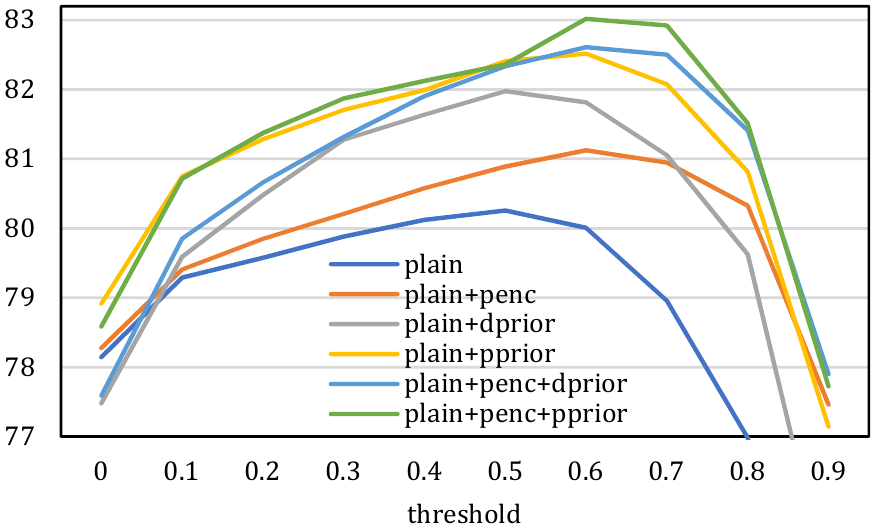}
	\caption{The matching F1 scores with respect to matching confidence of different network configurations on ETH3D dataset \cite{eth3d}.}
	\label{f1_cmp_eth}
\end{figure}

\begin{figure}[!t]
	\centering
	\includegraphics[]{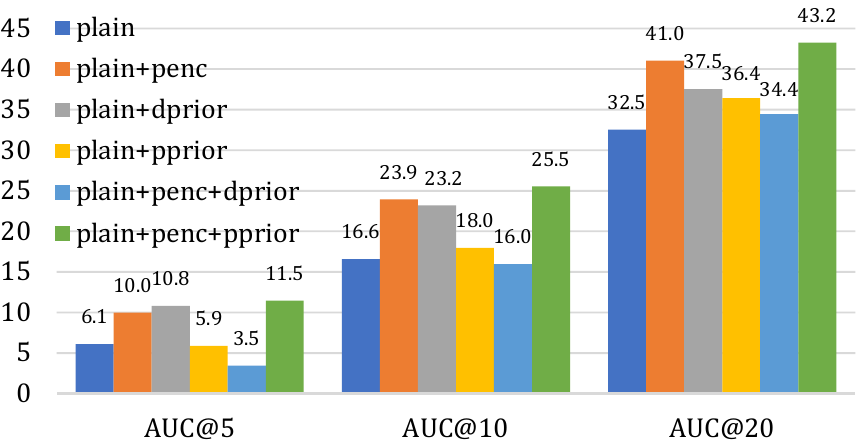}
	\caption{The pose estimation AUC of different network configurations on ETH3D dataset \cite{eth3d}.}
	\label{auc_cmp_eth}
\end{figure}

\subsubsection{Generalization to homographic image pairs}
To demonstrate the generalization of the proposed method, experiments are also carried out on homography image pairs. Table \ref{table:homo} illustrates the matching results of full SuperGlue \cite{superglue} (with ``outdoor" weights), SuperGlue of two attentional GNN layers, and the proposed method. Note that the latter two models were trained only on the RGBD datasets, no homography image pairs are included in the training process. As in Table \ref{table:homo}, the full SuperGlue \cite{superglue} with ``outdoor" weights achieves the best matching performance, giving F1 scores of 82.52\%, 82.73\%, and 75.51\% on HPatches illumination, viewpoint, and OxfordParis, respectively. However, F1 scores of the two GNN layers version of SuperGlue drops to 60.77\%, 51.12\%, and 45.58\% on these datasets. While the F1 scores of the proposed method significantly outperform the two GNN layers version of SuperGlue, which are 64.73\%, 62.18\%, and 51.42\% respectively. Moreover, as the viewpoint differences on HPatches viewpoint and OxfordParis are larger, the assistance of the spatial distribution prior is more significant. Therefore, the improvement of the F1 score is more obvious, which increases by 10.05\% and 5.84\% respectively. In addition to the matching performance, the homography estimation accuracy $Acc_H$ is also evaluated. The mean reprojection error of the four corners of the image is first computed, and then the homography estimation accuracy is defined by the accuracy of the corner error under a threshold of three pixels. As shown in Table \ref{table:homo}, the homography accuracy $Acc_H$ of the proposed method is slightly lower than SuperGlue \cite{superglue} on the Hpatches illumination dataset. The proposed method, on the other hand, significantly outperforms SueprGlue of two attentional GNN layers on HPatches viewpoint and OxfordParis datasets. The reason for this phenomenon is that for the Hpatches illumination dataset, the homography matrix is identity, and the noise added during generating the priors interferes with the matching process. For HPatches viewpoint and OxfordParis datasets, there are significant viewpoint differences in a pair of images. Thus the noise is trivial and the prior plays a vital role in improving the matching performance.

\begin{table}[!tb]		
	\centering 
	\caption{Ablation studies on ETH3D dataset \cite{eth3d}. The matching precision $P_m$, recall $R_m$ and $F1$ score are computed when the matching confidence is 0.2. The plain network contains two attentional GNN layers and an optimization layer, the ``penc" is the MLP position encoder in Equation (\ref{penc}), and the ``dprior" and ``pprior" denote the direct and probabilistic prior integration method respectively.}
	\begin{tabular}{ccccccc}
		\hline
		plain    &    penc    &   dprior   &   pprior   &     $P_m$      &     $R_m$      &      $F1$      \\ \hline
		\checkmark &            &            &            &     70.71      &     92.71      &     79.58      \\
		\checkmark & \checkmark &            &            &     70.73      &     93.35      &     79.85      \\
		\checkmark &            & \checkmark &            &     72.48      &     92.04      &     80.48      \\
		\checkmark &            &            & \checkmark & \textbf{73.29} &     93.09      &     81.28      \\
		\checkmark & \checkmark & \checkmark &            &     72.13      &     93.25      &     80.66      \\
		\checkmark & \checkmark &            & \checkmark &     72.94      & \textbf{93.95} & \textbf{81.37} \\ \hline
	\end{tabular}	
	\label{table:ablation}
\end{table}

\subsection{Ablation studies}
\label{ablation_sec}
To investigate the effectiveness of each model part, this subsection studies the position encoder, the direct and probabilistic prior integration, the matching loss and projection loss, as well as the number of attentional GNN layers to the matching performance.

\subsubsection{Position encoder}
The MLP in Equation (\ref{penc}) makes use of position by embedding it to feature space, while the proposed prior integration utilizes the position by propagating contextual keypoint features through the attentional GNN. The two methods of position utilization are compared in this section. As shown in Table \ref{table:ablation}, the position encoder (penc) promotes the F1 score on the plain network by 0.27\%. The proposed direct prior (dprior) and probabilistic prior integration boost the F1 score by 0.9\% and 2.7\% respectively. The same results can also be found in Fig. \ref{f1_cmp_eth}, where the F1 score is plotted with respect to the matching confidence. It could be because the embedding of positional encoding changes the feature distribution of descriptors, and it is difficult for the network to learn such embedding. While the proposed prior integration does not change the distribution of any features, but only aggregates the nearby features in the original feature space, so the proposed prior integration methods significantly improve the matching performance.

\subsubsection{Direct and probabilistic prior integration}
As shown in Table \ref{table:ablation}, compared to the plain network, after integrating the direct and probabilistic prior into attentional GNN, the precision is improved by 1.77\% and 2.58\% respectively, and the F1 score is improved 0.9\% and 1.7\% respectively. With the keypoint position encoder module, the total improvement of the F1 score is 1.08\% and 1.79\% for the direct and probabilistic prior integration respectively. The same promotion can also be found in Fig. \ref{f1_cmp_eth}. These primary matching results indicate the probabilistic prior integration is more efficient than direct prior integration.

In terms of the accuracy of the estimated pose, inconsistent results are observed for different approaches with the matching results as in Fig. \ref{auc_cmp_eth}. The improvements of AUCs by the direct prior integration are larger than those of probabilistic prior integration, and the improvements of both prior integrations are lower than the positional encoder except that of AUC under 5\textdegree. Another interesting result is that the combination of positional encoding and direct prior integration actually deteriorates the pose estimation accuracy. The possible reason could be that the positional embedding feature could not be processed correctly by the direct prior integration strategy. While the combination of positional encoding and probabilistic prior integration can promote each other, and improve the pose accuracy significantly.

\begin{figure}[!t]
	\centering
	\subfigure[]{\includegraphics[]{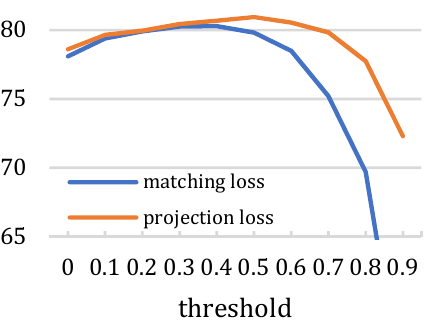}\label{losstype}}
	\subfigure[]{\includegraphics[]{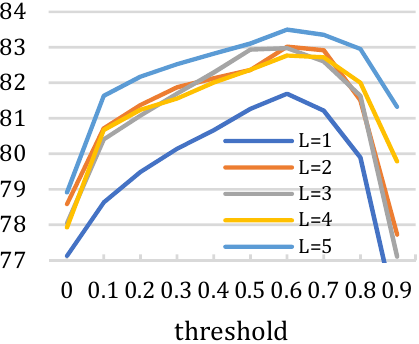}\label{layers}}
	\caption{Ablation studies of loss functions and the number of attentional GNN layers. (a): comparison of the matching loss and projection loss of F1 scores with respect to the matching confidence. (b): the matching F1 scores of the proposed model with different numbers of attentional GNN layers.}
\end{figure}

\subsubsection{Matching and projection loss}
\label{eval_loss}
We evaluated the matching loss of Equation (\ref{pos_neg}) and the projection loss of Equation (\ref{projection}) by training the plain network (with two attentional GNN layers and an optimization layer) on the same train dataset with a fixed random seed. As expected, the projection loss recalls more matches than matching loss due to its relaxation of the matching threshold, thus yields higher $F1$ scores as shown in Fig. \ref{losstype}. Thus, we use the projection loss for all experiments.

\subsubsection{Number of GNN layers}
\label{num_layer}
Since the introduction of the spatial distribution prior can assist the model to utilize the contextual features, so that the number of attentional GNN layers is reduced for efficiency. To verify the impact of the number of GNN layers on the performance, we trained and assessed the models of one to five attention GNN layers. The matching results are shown in Fig. \ref{layers}. Clearly, the model with more GNN layers has a better matching performance. However, when the number of GNN layers is larger than one, there is no significant performance improvement, but the running time is greatly increased. As shown in Table \ref{table:layers}, the pose estimation accuracy even decreases with the increase of the number of GNN layers, as a deeper model is harder to train. To balance the performance and efficiency, we use two attentional GNN layers.

\begin{table}[!t]		
	\centering 
	\caption{Matching performance for the different number of GNN layers on ETH3D \cite{eth3d} test set with matching confidence of 0.2.}
	\begin{tabular}{cccccc}
		\hline
		&  L=1  &  L=2  &  L=3  &  L=4  &  L=5  \\ \hline
		$Pm$    & 70.45 & 72.94 & 72.63 & 73.08 & 74.17 \\
		$Rm$    & 92.96 & 93.95 & 93.70 & 93.34 & 94.03 \\
		$F1$    & 79.49 & 81.37 & 81.07 & 81.24 & 82.17 \\
		AUC@10   & 24.03 & 25.53 & 18.92 & 24.03 & 21.85 \\ \hline
		time (ms) & 12.09 & 15.36 & 18.57 & 22.36 & 26.02 \\ \hline
	\end{tabular}	
	\label{table:layers}
\end{table}

\section{Conclusions}
\label{conclusion}

In this paper, motion prior from other sensors such as the IMU is adopted to obtain the spatial distribution prior of keypoints, which is exploited to streamline the attentional keypoint matching network. Specifically, the spatial distribution prior is naturally integrated into the attentional GNN network with the probabilistic perspective of attention. Thus, the number of GNN layers can be reduced, and the running time is decreased to about 15ms while keeping the matching performance. Besides, a loss using the pixel projection errors is proposed to train the network to achieve better matching performance. The experiments on SLAM datasets InteriorNet, TUM-RGBD, and ETH3D validate the effectiveness and efficiency of the proposed method. %Image processing plays an important role in autonomous systems. 
A similar idea can be adopted to study other image processing problems such as motion de-blurring, visual tracking for autonomous systems. All these problems will be studied in our future research. 

\section*{Acknowledgment}
This work was supported by the National Nature Science Foundation of China under grant 61620106012.

\ifCLASSOPTIONcaptionsoff
\newpage
\fi

\bibliographystyle{IEEEtran.bst}
%\bibliography{bibliography.bib}

\begin{thebibliography}{10}
	\providecommand{\url}[1]{#1}
	\csname url@samestyle\endcsname
	\providecommand{\newblock}{\relax}
	\providecommand{\bibinfo}[2]{#2}
	\providecommand{\BIBentrySTDinterwordspacing}{\spaceskip=0pt\relax}
	\providecommand{\BIBentryALTinterwordstretchfactor}{4}
	\providecommand{\BIBentryALTinterwordspacing}{\spaceskip=\fontdimen2\font plus
		\BIBentryALTinterwordstretchfactor\fontdimen3\font minus
		\fontdimen4\font\relax}
	\providecommand{\BIBforeignlanguage}[2]{{%
			\expandafter\ifx\csname l@#1\endcsname\relax
			\typeout{** WARNING: IEEEtran.bst: No hyphenation pattern has been}%
			\typeout{** loaded for the language `#1'. Using the pattern for}%
			\typeout{** the default language instead.}%
			\else
			\language=\csname l@#1\endcsname
			\fi
			#2}}
	\providecommand{\BIBdecl}{\relax}
	\BIBdecl
	
	\bibitem{harris}
	C.~G. Harris and M.~Stephens, ``A combined corner and edge detector.'' in
	\emph{Alvey vision conference}, vol.~15.\hskip 1em plus 0.5em minus
	0.4em\relax Citeseer, 1988, pp. 10--5244.
	
	\bibitem{shi}
	J.~Shi, ``Good features to track,'' in \emph{1994 {Proceedings} of {IEEE}
		conference on computer vision and pattern recognition}.\hskip 1em plus 0.5em
	minus 0.4em\relax IEEE, 1994, pp. 593--600.
	
	\bibitem{fast}
	S.~Alkaabi and F.~Deravi, ``Candidate pruning for fast corner detection,''
	\emph{Electronics Letters}, vol.~40, no.~1, pp. 18--19, 2004.
	
	\bibitem{keynet}
	A.~Barroso-Laguna, E.~Riba, D.~Ponsa, and K.~Mikolajczyk, ``Key. net:
	{Keypoint} detection by handcrafted and learned cnn filters,'' in
	\emph{Proceedings of the {IEEE} {International} {Conference} on {Computer}
		{Vision}}, 2019, pp. 5836--5844.
	
	\bibitem{hartley_multiple_2003}
	R.~Hartley and A.~Zisserman, \emph{Multiple view geometry in computer
		vision}.\hskip 1em plus 0.5em minus 0.4em\relax Cambridge university press,
	2003.
	
	\bibitem{ding_multi-camera_2020}
	C.~Ding and Z.~Ma, ``Multi-{Camera} {Color} {Correction} via {Hybrid}
	{Histogram} {Matching},'' \emph{IEEE Transactions on Circuits and Systems for
		Video Technology}, Dec. 2020, doi: 10.1109/TCSVT.2020.3038484.
	
	\bibitem{wang_detail-enhanced_2020}
	Q.~Wang, W.~Chen, X.~Wu, and Z.~Li, ``Detail-enhanced multi-scale exposure
	fusion in {YUV} color space,'' \emph{IEEE Transactions on Circuits and
		Systems for Video Technology}, vol.~30, no.~8, pp. 2418--2429, Aug. 2020.
	
	\bibitem{zheng_single_2020}
	C.~Zheng, Z.~Li, Y.~Yang, and S.~Wu, ``Single {Image} {Brightening} via
	{Multi}-{Scale} {Exposure} {Fusion} with {Hybrid} {Learning},'' \emph{IEEE
		Transactions on Circuits and Systems for Video Technology}, Jun. 2020, doi:
	10.1109/TCSVT.2020.3009235.
	
	\bibitem{ma_locality_2019}
	J.~Ma, J.~Zhao, J.~Jiang, H.~Zhou, and X.~Guo,
	``\BIBforeignlanguage{en}{Locality {Preserving} {Matching}},''
	\emph{\BIBforeignlanguage{en}{International Journal of Computer Vision}},
	vol. 127, no.~5, pp. 512--531, May 2019.
	
	\bibitem{sift}
	D.~G. Lowe, ``Distinctive image features from scale-invariant keypoints,''
	\emph{International journal of computer vision}, vol.~60, no.~2, pp. 91--110,
	2004.
	
	\bibitem{pele_linear_2008}
	O.~Pele and M.~Werman, ``A linear time histogram metric for improved sift
	matching,'' in \emph{European conference on computer vision}.\hskip 1em plus
	0.5em minus 0.4em\relax Springer, 2008, pp. 495--508.
	
	\bibitem{li_rejecting_2010}
	X.~Li and Z.~Hu, ``Rejecting mismatches by correspondence function,''
	\emph{International Journal of Computer Vision}, vol.~89, no.~1, pp. 1--17,
	2010.
	
	\bibitem{lipman_feature_2014}
	Y.~Lipman, S.~Yagev, R.~Poranne, D.~W. Jacobs, and R.~Basri, ``Feature matching
	with bounded distortion,'' \emph{ACM Transactions on Graphics (TOG)},
	vol.~33, no.~3, pp. 1--14, 2014.
	
	\bibitem{ma_robust_2014}
	J.~Ma, J.~Zhao, J.~Tian, A.~L. Yuille, and Z.~Tu,
	``\BIBforeignlanguage{en}{Robust {Point} {Matching} via {Vector} {Field}
		{Consensus}},'' \emph{\BIBforeignlanguage{en}{IEEE Transactions on Image
			Processing}}, vol.~23, no.~4, pp. 1706--1721, Apr. 2014.
	
	\bibitem{lin_code_2017}
	W.-Y. Lin, F.~Wang, M.-M. Cheng, S.-K. Yeung, P.~H. Torr, M.~N. Do, and J.~Lu,
	``{CODE}: {Coherence} based decision boundaries for feature correspondence,''
	\emph{IEEE transactions on pattern analysis and machine intelligence},
	vol.~40, no.~1, pp. 34--47, 2017.
	
	\bibitem{pointcn}
	K.~M. Yi, E.~Trulls, Y.~Ono, V.~Lepetit, M.~Salzmann, and P.~Fua, ``Learning to
	{Find} {Good} {Correspondences},'' in \emph{2018 {IEEE}/{CVF} {Conference} on
		{Computer} {Vision} and {Pattern} {Recognition}}.\hskip 1em plus 0.5em minus
	0.4em\relax Salt Lake City, UT, USA: IEEE, Jun. 2018, pp. 2666--2674.
	
	\bibitem{nmnet}
	C.~Zhao, Z.~Cao, C.~Li, X.~Li, and J.~Yang, ``{NM}-{Net}: {Mining} {Reliable}
	{Neighbors} for {Robust} {Feature} {Correspondences},'' in \emph{Proceedings
		of the {IEEE} {Conference} on {Computer} {Vision} and {Pattern}
		{Recognition}}, 2018.
	
	\bibitem{oanet}
	J.~Zhang, D.~Sun, Z.~Luo, A.~Yao, L.~Zhou, T.~Shen, Y.~Chen, H.~Liao, and
	L.~Quan, ``Learning {Two}-{View} {Correspondences} and {Geometry} {Using}
	{Order}-{Aware} {Network},'' in \emph{2019 {IEEE}/{CVF} {International}
		{Conference} on {Computer} {Vision} ({ICCV})}.\hskip 1em plus 0.5em minus
	0.4em\relax Seoul, Korea (South): IEEE, Oct. 2019, pp. 5844--5853.
	
	\bibitem{sun_acne_2020}
	W.~Sun, W.~Jiang, E.~Trulls, A.~Tagliasacchi, and K.~M. Yi, ``{ACNe}:
	{Attentive} {Context} {Normalization} for {Robust}
	{Permutation}-{Equivariant} {Learning},'' in \emph{Proceedings of the
		{IEEE}/{CVF} {Conference} on {Computer} {Vision} and {Pattern}
		{Recognition}}, 2020, pp. 11\,286--11\,295.
	
	\bibitem{surf}
	H.~Bay, T.~Tuytelaars, and L.~Van~Gool, ``Surf: {Speeded} up robust features,''
	in \emph{European conference on computer vision}.\hskip 1em plus 0.5em minus
	0.4em\relax Springer, 2006, pp. 404--417.
	
	\bibitem{lift}
	K.~M. Yi, E.~Trulls, V.~Lepetit, and P.~Fua, ``{LIFT}: {Learned} {Invariant}
	{Feature} {Transform},'' in \emph{European {Conference} on {Computer}
		{Vision}}, vol. 9910.\hskip 1em plus 0.5em minus 0.4em\relax Cham: Springer,
	2016, pp. 467--483.
	
	\bibitem{matchnet}
	X.~Han, T.~Leung, Y.~Jia, R.~Sukthankar, and A.~C. Berg, ``Matchnet: {Unifying}
	feature and metric learning for patch-based matching,'' in \emph{Proceedings
		of the {IEEE} {Conference} on {Computer} {Vision} and {Pattern}
		{Recognition}}, 2015, pp. 3279--3286.
	
	\bibitem{hardnet}
	A.~Mishchuk, D.~Mishkin, F.~Radenovic, and J.~Matas,
	``\BIBforeignlanguage{en}{Working hard to know your neighbor's margins:
		{Local} descriptor learning loss},'' in
	\emph{\BIBforeignlanguage{en}{Advances in {Neural} {Information} {Processing}
			{Systems}}}, Jan. 2018.
	
	\bibitem{sosnet}
	Y.~Tian, X.~Yu, B.~Fan, F.~Wu, H.~Heijnen, and V.~Balntas, ``{SOSNet}: {Second}
	{Order} {Similarity} {Regularization} for {Local} {Descriptor} {Learning},''
	in \emph{Conference on {Computer} {Vision} and {Pattern} {Recognition}}, Dec.
	2019.
	
	\bibitem{superpoint}
	D.~DeTone, T.~Malisiewicz, and A.~Rabinovich, ``{SuperPoint}:
	{Self}-{Supervised} {Interest} {Point} {Detection} and {Description},'' in
	\emph{Proceedings of the {IEEE} {Conference} on {Computer} {Vision} and
		{Pattern} {Recognition} {Workshops}}, 2018, pp. 224--236.
	
	\bibitem{d2net}
	M.~Dusmanu, I.~Rocco, T.~Pajdla, M.~Pollefeys, J.~Sivic, A.~Torii, and
	T.~Sattler, ``D2-{Net}: {A} {Trainable} {CNN} for {Joint} {Description} and
	{Detection} of {Local} {Features},'' in \emph{2019 {IEEE}/{CVF} {Conference}
		on {Computer} {Vision} and {Pattern} {Recognition} ({CVPR})}.\hskip 1em plus
	0.5em minus 0.4em\relax Long Beach, CA, USA: IEEE, Jun. 2019, pp. 8084--8093.
	
	\bibitem{r2d2}
	J.~Revaud, P.~Weinzaepfel, C.~D. Souza, N.~Pion, G.~Csurka, Y.~Cabon, and
	M.~Humenberger, ``{R2D2}: {Repeatable} and {Reliable} {Detector} and
	{Descriptor},'' in \emph{{NeurIPS}}, 2019, p.~12.
	
	\bibitem{superglue}
	P.-E. Sarlin, D.~DeTone, T.~Malisiewicz, and A.~Rabinovich, ``{SuperGlue}:
	{Learning} {Feature} {Matching} with {Graph} {Neural} {Networks},'' in
	\emph{Proceedings of the {IEEE}/{CVF} {Conference} on {Computer} {Vision} and
		{Pattern} {Recognition}}, Mar. 2020, pp. 4938--4947.
	
	\bibitem{orbslam2}
	R.~Mur-Artal and J.~D. Tardós, ``{ORB}-{SLAM2}: {An} open-source slam system
	for monocular, stereo, and rgb-d cameras,'' \emph{IEEE Transactions on
		Robotics}, vol.~33, no.~5, pp. 1255--1262, 2017.
	
	\bibitem{loam}
	J.~Zhang and S.~Singh, ``{LOAM}: {Lidar} {Odometry} and {Mapping} in
	{Real}-time.''\hskip 1em plus 0.5em minus 0.4em\relax Robotics: Science and
	Systems Foundation, Jul. 2014.
	
	\bibitem{accimu}
	J.~Henawy, Z.~Li, W.~Y. Yau, G.~Seet, and K.~W. Wan, ``Accurate {IMU}
	{Preintegration} {Using} {Switched} {Linear} {Systems} {For} {Autonomous}
	{Systems},'' in \emph{2019 {IEEE} {Intelligent} {Transportation} {Systems}
		{Conference} ({ITSC})}.\hskip 1em plus 0.5em minus 0.4em\relax Auckland, New
	Zealand: IEEE, Oct. 2019, pp. 3839--3844.
	
	\bibitem{accimu2}
	J.~Henawy, Z.~Li, W.~Y. Yau, and G.~Seet, ``Accurate {IMU} {Factor} {Using}
	{Switched} {Linear} {Systems} {For} {VIO},'' \emph{IEEE Transactions on
		Industrial Electronics}, vol.~68, no.~9, pp. 1--10, Sep. 2021.
	
	\bibitem{interiornet_2018}
	W.~Li, S.~Saeedi, J.~McCormac, R.~Clark, D.~Tzoumanikas, Q.~Ye, Y.~Huang,
	R.~Tang, and S.~Leutenegger, ``{InteriorNet}: {Mega}-scale {Multi}-sensor
	{Photo}-realistic {Indoor} {Scenes} {Dataset},'' in \emph{British {Machine}
		{Vision} {Conference} ({BMVC})}, 2018.
	
	\bibitem{tumrgbd}
	J.~Sturm, N.~Engelhard, F.~Endres, W.~Burgard, and D.~Cremers, ``A benchmark
	for the evaluation of {RGB}-{D} {SLAM} systems,'' in \emph{2012 {IEEE}/{RSJ}
		{International} {Conference} on {Intelligent} {Robots} and {Systems}}, Oct.
	2012, pp. 573--580.
	
	\bibitem{eth3d}
	T.~Schops, T.~Sattler, and M.~Pollefeys, ``{BAD} {SLAM}: {Bundle} {Adjusted}
	{Direct} {RGB}-{D} {SLAM},'' in \emph{2019 {IEEE}/{CVF} {Conference} on
		{Computer} {Vision} and {Pattern} {Recognition} ({CVPR})}.\hskip 1em plus
	0.5em minus 0.4em\relax Long Beach, CA, USA: IEEE, Jun. 2019, pp. 134--144.
	
	\bibitem{orb}
	E.~Rublee, V.~Rabaud, K.~Konolige, and G.~Bradski, ``{ORB}: {An} efficient
	alternative to {SIFT} or {SURF},'' in \emph{2011 {International} conference
		on computer vision}.\hskip 1em plus 0.5em minus 0.4em\relax Ieee, 2011, pp.
	2564--2571.
	
	\bibitem{tfeat}
	V.~Balntas, E.~Riba, D.~Ponsa, and K.~Mikolajczyk, ``Learning local feature
	descriptors with triplets and shallow convolutional neural networks.'' in
	\emph{{BMVC}}, vol.~1, 2016, p.~3.
	
	\bibitem{l2net}
	Y.~Tian, B.~Fan, and F.~Wu, ``\BIBforeignlanguage{en}{L2-{Net}: {Deep}
		{Learning} of {Discriminative} {Patch} {Descriptor} in {Euclidean}
		{Space}},'' in \emph{\BIBforeignlanguage{en}{2017 IEEE Conference on Computer
			Vision and Pattern Recognition}}.\hskip 1em plus 0.5em minus 0.4em\relax
	Honolulu, HI: IEEE, Jul. 2017, pp. 6128--6136.
	
	\bibitem{lfnet}
	Y.~Ono, E.~Trulls, P.~Fua, and K.~M. Yi, ``{LF}-{Net}: {Learning} {Local}
	{Features} from {Images},'' in \emph{Advances in {Neural} {Information}
		{Processing} {Systems} 31}.\hskip 1em plus 0.5em minus 0.4em\relax Curran
	Associates, Inc., 2018, pp. 6234--6244.
	
	\bibitem{tilde}
	Y.~Verdie, K.~Yi, P.~Fua, and V.~Lepetit, ``Tilde: {A} temporally invariant
	learned detector,'' in \emph{Proceedings of the {IEEE} {Conference} on
		{Computer} {Vision} and {Pattern} {Recognition}}, 2015, pp. 5279--5288.
	
	\bibitem{quadnet}
	N.~Savinov, A.~Seki, L.~Ladicky, T.~Sattler, and M.~Pollefeys, ``Quad-networks:
	unsupervised learning to rank for interest point detection,'' in
	\emph{Proceedings of the {IEEE} conference on computer vision and pattern
		recognition}, 2017, pp. 1822--1830.
	
	\bibitem{guo_good_2012}
	X.~Guo and X.~Cao, ``Good match exploration using triangle constraint,''
	\emph{Pattern Recognition Letters}, vol.~33, no.~7, pp. 872--881, 2012.
	
	\bibitem{ma_guided_2018}
	J.~Ma, J.~Jiang, H.~Zhou, J.~Zhao, and X.~Guo, ``Guided locality preserving
	feature matching for remote sensing image registration,'' \emph{IEEE
		transactions on geoscience and remote sensing}, vol.~56, no.~8, pp.
	4435--4447, 2018.
	
	\bibitem{gms}
	J.~Bian, W.-Y. Lin, Y.~Matsushita, S.-K. Yeung, T.-D. Nguyen, and M.-M. Cheng,
	``Gms: {Grid}-based motion statistics for fast, ultra-robust feature
	correspondence,'' in \emph{Proceedings of the {IEEE} {Conference} on
		{Computer} {Vision} and {Pattern} {Recognition}}, 2017, pp. 4181--4190.
	
	\bibitem{hu_matching_2015}
	Y.-T. Hu, Y.-Y. Lin, H.-Y. Chen, K.-J. Hsu, and B.-Y. Chen, ``Matching images
	with multiple descriptors: {An} unsupervised approach for locally adaptive
	descriptor selection,'' \emph{IEEE Transactions on Image Processing},
	vol.~24, no.~12, pp. 5995--6010, 2015.
	
	\bibitem{maier_guided_2016}
	J.~Maier, M.~Humenberger, M.~Murschitz, O.~Zendel, and M.~Vincze, ``Guided
	matching based on statistical optical flow for fast and robust correspondence
	analysis,'' in \emph{European {Conference} on {Computer} {Vision}}.\hskip 1em
	plus 0.5em minus 0.4em\relax Springer, 2016, pp. 101--117.
	
	\bibitem{liu_regularization_2015}
	Y.~Liu, L.~De~Dominicis, B.~Wei, L.~Chen, and R.~R. Martin, ``Regularization
	based iterative point match weighting for accurate rigid transformation
	estimation,'' \emph{IEEE transactions on visualization and computer
		graphics}, vol.~21, no.~9, pp. 1058--1071, 2015.
	
	\bibitem{torresani_feature_2008}
	L.~Torresani, V.~Kolmogorov, and C.~Rother, ``Feature {Correspondence} via
	{Graph} {Matching}: {Models} and {Global} {Optimization},'' in
	\emph{{European} {Conference} on {Computer} {Vision}}, 2008.
	
	\bibitem{bbmatch}
	S.~{Liu}, H.~{Wang}, Y.~{Wei}, and C.~{Pan}, ``Bb-homography: Joint binary
	features and bipartite graph matching for homography estimation,'' \emph{IEEE
		Transactions on Circuits and Systems for Video Technology}, vol.~25, no.~2,
	pp. 239--250, 2015.
	
	\bibitem{coor_gm}
	R.~{Zhang} and W.~{Wang}, ``Second- and high-order graph matching for
	correspondence problems,'' \emph{IEEE Transactions on Circuits and Systems
		for Video Technology}, vol.~28, no.~10, pp. 2978--2992, 2018.
	
	\bibitem{dcm}
	Y.~F. {Yu}, G.~{Xu}, K.~K. {Huang}, H.~{Zhu}, L.~{Chen}, and H.~{Wang}, ``Dual
	calibration mechanism based l2,p-norm for graph matching,'' \emph{IEEE
		Transactions on Circuits and Systems for Video Technology}, 2020,
	doi:10.1109/TCSVT.2020.3023781.
	
	\bibitem{ransac}
	M.~A. Fischler and R.~C. Bolles, ``Random sample consensus: a paradigm for
	model fitting with applications to image analysis and automated
	cartography,'' \emph{Communications of the ACM}, vol.~24, no.~6, pp.
	381--395, 1981.
	
	\bibitem{ufer_deep_2017}
	N.~Ufer and B.~Ommer, ``Deep semantic feature matching,'' in \emph{Proceedings
		of the {IEEE} {Conference} on {Computer} {Vision} and {Pattern}
		{Recognition}}, 2017, pp. 6914--6923.
	
	\bibitem{yu_hierarchical_2018}
	W.~Yu, X.~Sun, K.~Yang, Y.~Rui, and H.~Yao, ``Hierarchical semantic image
	matching using {CNN} feature pyramid,'' \emph{Computer Vision and Image
		Understanding}, vol. 169, pp. 40--51, 2018.
	
	\bibitem{cur}
	S.~{Khan}, M.~{Nawaz}, X.~{Guoxia}, and H.~{Yan}, ``Image correspondence with
	cur decomposition-based graph completion and matching,'' \emph{IEEE
		Transactions on Circuits and Systems for Video Technology}, vol.~30, no.~9,
	pp. 3054--3067, 2020.
	
	\bibitem{1kou2017}
	F.~{Kou}, Z.~{Li}, C.~{Wen}, and W.~{Chen}, ``Multi-scale exposure fusion via
	gradient domain guided image filtering,'' in \emph{2017 IEEE International
		Conference on Multimedia and Expo (ICME)}, 2017, pp. 1105--1110.
	
	\bibitem{1zheng2013}
	J.~{Zheng}, Z.~{Li}, Z.~{Zhu}, S.~{Wu}, and S.~{Rahardja}, ``Hybrid patching
	for a sequence of differently exposed images with moving objects,''
	\emph{IEEE Transactions on Image Processing}, vol.~22, no.~12, pp.
	5190--5201, 2013.
	
	\bibitem{pointnet}
	C.~R. Qi, H.~Su, K.~Mo, and L.~J. Guibas, ``{PointNet}: {Deep} {Learning} on
	{Point} {Sets} for {3D} {Classification} and {Segmentation},'' in
	\emph{Proceedings of the {IEEE} conference on computer vision and pattern
		recognition}, 2017, pp. 652--660.
	
	\bibitem{sinkhorn}
	M.~Cuturi, ``Sinkhorn {Distances}: {Lightspeed} {Computation} of {Optimal}
	{Transport},'' in \emph{Advances in {Neural} {Information} {Processing}
		{Systems} 26}.\hskip 1em plus 0.5em minus 0.4em\relax Curran Associates,
	Inc., 2013, pp. 2292--2300.
	
	\bibitem{viorb}
	R.~Mur-Artal and J.~D. Tardós, ``Visual-inertial monocular {SLAM} with map
	reuse,'' \emph{IEEE Robotics and Automation Letters}, vol.~2, no.~2, pp.
	796--803, 2017.
	
	\bibitem{vins}
	T.~Qin, P.~Li, and S.~Shen, ``{VINS}-{Mono}: {A} {Robust} and {Versatile}
	{Monocular} {Visual}-{Inertial} {State} {Estimator},'' \emph{IEEE
		Transactions on Robotics}, vol.~34, no.~4, pp. 1004--1020, Aug. 2018.
	
	\bibitem{manifoldimu}
	C.~Forster, L.~Carlone, F.~Dellaert, and D.~Scaramuzza, ``On-{Manifold}
	{Preintegration} for {Real}-{Time} {Visual}–{Inertial} {Odometry},''
	\emph{IEEE Transactions on Robotics}, vol.~33, no.~1, pp. 1--21, 2017.
	
	\bibitem{atten_transformer}
	A.~Vaswani, N.~Shazeer, N.~Parmar, J.~Uszkoreit, L.~Jones, A.~N. Gomez,
	L.~Kaiser, and I.~Polosukhin, ``Attention is {All} you {Need},'' in
	\emph{Advances in {Neural} {Information} {Processing} {Systems} 30}.\hskip
	1em plus 0.5em minus 0.4em\relax Curran Associates, Inc., 2017, pp.
	5998--6008.
	
	\bibitem{wang2018non}
	X.~Wang, R.~Girshick, A.~Gupta, and K.~He, ``Non-local neural networks,'' in
	\emph{Proceedings of the IEEE conference on computer vision and pattern
		recognition}, 2018, pp. 7794--7803.
	
	\bibitem{optimal_transport}
	G.~Peyré and M.~Cuturi, ``Computational {Optimal} {Transport},''
	\emph{arXiv:1803.00567 [stat]}, Mar. 2020.
	
	\bibitem{munkres_algorithms_1957}
	J.~Munkres, ``Algorithms for the assignment and transportation problems,''
	\emph{Journal of the society for industrial and applied mathematics}, vol.~5,
	no.~1, pp. 32--38, 1957.
	
	\bibitem{hpatches_2017}
	V.~Balntas, K.~Lenc, A.~Vedaldi, and K.~Mikolajczyk, ``{HPatches}: {A}
	benchmark and evaluation of handcrafted and learned local descriptors,'' in
	\emph{Proceedings of the {IEEE} {Conference} on {Computer} {Vision} and
		{Pattern} {Recognition}}, 2017, pp. 5173--5182.
	
	\bibitem{paris}
	F.~Radenović, A.~Iscen, G.~Tolias, Y.~Avrithis, and O.~Chum, ``Revisiting
	oxford and paris: {Large}-scale image retrieval benchmarking,'' in
	\emph{Proceedings of the {IEEE} {Conference} on {Computer} {Vision} and
		{Pattern} {Recognition}}, 2018, pp. 5706--5715.
	
	\bibitem{adam}
	D.~P. Kingma and J.~Ba, ``Adam: {A} {Method} for {Stochastic} {Optimization},''
	in \emph{ICLR}, 2015.
	
\end{thebibliography}

% Generated by IEEEtran.bst, version: 1.14 (2015/08/26)

\vfill

\end{document}